\documentclass[pmlr]{jmlr}

\RequirePackage{graphicx}
\usepackage{booktabs}
\usepackage{caption}
\usepackage{longtable}
\usepackage{booktabs}
\usepackage{siunitx}
\usepackage{multirow}
\usepackage{dashrule}
\usepackage{float}
\usepackage{hyperref}

\makeatletter
\def\set@curr@file#1{\def\@curr@file{#1}}
\makeatother
\usepackage{siunitx}
\usepackage{arydshln}

\jmlrproceedings{PMLR}{Proceedings of Machine Learning Research}
\jmlrvolume{340}
\jmlryear{2026}
\jmlrworkshop{Machine Learning for Healthcare}

\title[PICU stewardship benchmarking]{Benchmarking Machine Learning Architectures for Antimicrobial Stewardship in Pediatric ICUs}

\author{%
  \Name{Niklas Raehse$^{1,2}$} \Email{niklas.raehse@hest.ethz.ch}\\
  \Name{Luregn J. Schlapbach$^{1}$} \Email{luregn.schlapbach@kispi.uzh.ch}\\
  \Name{Daphné Chopard$^{1,3}$} \Email{daphne.chopard@kispi.uzh.ch}\\ 
  \addr $^{1}$Department of Intensive Care and Neonatology and Children’s Research Center, University of Zurich,
University Children’s Hospital Zurich, Zurich, Switzerland\\
  \addr $^{2}$Department of Health Sciences and Technology, ETH Zurich, Zurich, Switzerland\\
\addr $^{3}$Department of Computer Science, ETH Zurich, Zurich, Switzerland
}

\begin{document}

\maketitle

\begin{abstract}
Antimicrobial stewardship (AMS) is critical for reducing unnecessary antibiotic exposure, particularly in pediatric intensive care units (PICUs), where clinical uncertainty leads to frequent broad-spectrum use, and overuse amplifies antimicrobial resistance which have long-term consequences. Machine learning has been proposed to support AMS by identifying patient-level opportunities for intervention from electronic health record data. However, prior work focuses on adult populations and predominantly relies on static tabular representations, leaving open questions about target design, temporal modeling, and generalizability in pediatric settings.
In this work, we present a systematic benchmarking study of AMS intervention prediction in the PICU. Using the publicly available Paediatric Intensive Care database from China and a private PICU cohort from the University Children's Hospital Zurich, Switzerland, we define four clinically relevant proxy targets for reducing antibiotic use: intravenous-to-oral switching, de-escalation, discontinuation, and short-course therapy. We then compare tabular, sequence-based, and graph-based temporal models under a unified evaluation framework.
We find that model performance is primarily driven by target prevalence and data characteristics rather than model complexity. Sequence models provide improvements in precision-recall trade-off over tabular approaches at coarse (24-hour) resolution, with limited additional gains when finer temporal structure is incorporated. However, these gains come at the cost of poorer calibration, with simpler tabular models producing more reliable probability estimates. 
Our results highlight the importance of target selection, temporal representation, and calibration in clinical machine learning, and provide practical guidance for developing reliable decision support systems for AMS in pediatric critical care. The code is publicly available\footnote{\url{https://github.com/pinnacle-kispi/AMS_intervention_prediction}}.
\end{abstract}

\section{Introduction}


Antimicrobial resistance is a growing global threat, and optimizing antibiotic use through antimicrobial stewardship (AMS) is essential to mitigate its impact \citep{who2015gap, barlam2016implementing}. 
In the pediatric intensive care unit (PICU), antibiotics are among the most frequently administered therapies \citep{willemsOptimizingUseAntibiotic2021} and are often initiated by default due to non-specific signs of systemic inflammation in critically ill children  \citep{weissSurvivingSepsisCampaign2020, evansAssociationNewYork2018}. This leads to substantial overuse, much of which may be unnecessary or overly broad \citep{chiotos2022picu, abbas2016picu}, contributing to toxicity and antimicrobial resistance, with amplified long-term consequences in children \citep{romandini2021pediatric,neuman2018antibiotics,sivasankar2023variation,fanelliImprovingQualityHospital2020}. 
AMS programs aim to identify opportunities to modify therapy, including de-escalation, discontinuation, or switching from intravenous to oral treatment \citep{pollackCoreElementsHospital2014,khdourImpactAntimicrobialStewardship2018, deresinskiPrinciplesAntibioticTherapy2007}. 
However, these processes are resource-intensive and may miss timely intervention opportunities~\citep{who2019ams_toolkit}, motivating the development of a clinical decision support system (CDSS) to assist clinicians in identifying when antibiotic therapy could be safely adjusted~\citep{giacobbe2024explainable}. 
Machine learning (ML) methods using electronic health record (EHR) data have been proposed to support AMS by predicting when stewardship interventions are  ~\citep{giacobbe2024explainable, tran2024scaling}. These approaches typically rely on retrospective targets derived from historical clinical decisions, which reflect observed practice rather than optimal care, but remain useful for capturing real-world stewardship behavior at scale. Identifying contexts in which therapy modification is likely to be appropriate is particularly valuable, as clinicians may be hesitant to adjust treatment under uncertainty \citep{chiotosAntibioticsHowCan2020}.
Despite these advances, AMS intervention prediction in the PICU remains an open challenge \citep{fanelliRoleArtificialIntelligence2020}. Compared to adult settings, pediatric data are more heterogeneous \citep{brunsAdministrativeDataPediatric2022}, and little work has addressed stewardship prediction in this population. Consequently, it remains unclear which intervention targets are most informative and which modeling approaches best capture the temporal and clinical structure of stewardship decisions. Commonly used targets such as intravenous-to-oral switching are underutilized in pediatric settings \citep{berthe2012antibiotic, mcmullan2016ivoral_systematic}, highlighting the need to reassess target definitions.
From a modeling perspective, patient data can be represented as either static snapshots or temporal sequences, but the impact of these approaches has not been systematically studied in AMS literature, and model selection is often driven by convention rather than evidence.
In this work, we address these gaps by systematically benchmarking ML architectures for AMS intervention prediction in the PICU. We evaluate models across two cohorts, a publicly available PICU dataset \citep{zeng2020pic} and a private dataset, to obtain a broader understanding of findings across pediatric populations. We define four stewardship proxy outcomes clinically relevant to the PICU setting: intravenous-to-oral switching, de-escalation, discontinuation, and short-course therapy. We compare tabular, sequence-based, and graph-based temporal models under a consistent framework, providing insights into how temporal representation and model choice affect predictive performance and offering guidance for the development of a reliable CDSS for AMS in pediatric critical care.

\vspace{-0.2cm}
\subsection*{Generalizable Insights about Machine Learning in the Context of Healthcare}

Our findings highlight two broader insights for applying machine learning in clinical settings beyond antimicrobial stewardship. First, evaluation metrics can lead to different conclusions about model utility: similar AUROC values  can obscure meaningful differences in F1 score and AUPRC, especially for rare outcomes. Metrics should therefore be selected according to the intended clinical use, target prevalence, and relative costs of false-positive and false-negative predictions, and should include prevalence-sensitive metrics that better reflect a model's ability to identify clinically relevant positive cases. Second, prediction targets developed in well-studied populations may be clinically irrelevant or too rare in underrepresented groups; for example, intravenous-to-oral antibiotic switching is common in adults but arises less often in pediatric care. Clinical machine-learning studies should therefore develop and validate both models and targets specifically for populations whose care processes and outcome distributions differ from those represented in existing research.

\section{Related Work}
\paragraph{AMS intervention prediction targets.}
Several studies have proposed predicting AMS interventions from EHR data \citep{beaudoin2014ams, goodman2022paf, bystritsky2020asp, tran2024scaling}, operationalizing stewardship actions such as intravenous-to-oral switching, de-escalation, and discontinuation as prediction targets. 
However, most work focuses on a single intervention type~\citep{bolton2024personalising,bolton2025impact} or collapses multiple actions into a single binary outcome \citep{bystritsky2020asp,goodman2022paf}, thereby potentially obscuring differences between stewardship decisions. 
Furthermore, existing studies are primarily conducted in on adult populations, and it remains unclear how well these targets transfer to pediatric settings, where intervention patterns and data characteristics differ~\citep{fay2020antimicrobial, shahUseMachineLearning2023}.

\paragraph{Modeling approaches for AMS prediction.}

Prior work spans a range of modeling approaches. Early studies rely on tabular representations with classical models such as logistic regression or gradient boosting \citep{beaudoin2014ams, tran2024scaling}, prioritizing interpretability and ease of deployment. These approaches 
represent patients as snapshots by aggregating dynamic features into summary statistics, thereby discarding temporal structure for simplicity~\citep{beaudoin2014ams, goodman2022paf, bystritsky2020asp, tran2024scaling}. 
More recent work has explored deep neural networks with post-hoc explanation \citep{bolton2024personalising}, but these models operate on similar aggregated inputs and therefore do not explicitly model temporal dynamics. While sequence-based models can capture longitudinal information, they have not been systematically evaluated for AMS prediction. Existing studies remain fragmented across private and public datasets, limiting reproducibility and cross-site comparison, and do not jointly assess the impact of temporal representations and model choice.

\paragraph{Causal outcome estimation for AMS.}
A related line of work estimates outcomes under alternative treatment decisions as opposed to predicting observed interventions. For example, \cite{bolton2022ml_cessation} use a recurrent neural network and synthetic controls to model counterfactual antibiotic cessation. While this approach targets decision support more directly, it relies on strong assumptions and unobserved confounding. In contrast, predicting observed interventions provides a simpler and more reproducible supervised learning setting, and prior work has shown that such predictions can improve the prioritization of cases for stewardship review \citep{tran2024scaling}. However, as observed interventions remain proxies for optimal decisions, prospective evaluation is required to determine whether model-guided review improves stewardship efficiency and patient outcomes.

\paragraph{PICU-specific challenges.}
ML for pediatric AMS remains underexplored~\citep{fanelliRoleArtificialIntelligence2020}. PICU data often have high missingness likely due to the heterogenous casemix \citep{brunsAdministrativeDataPediatric2022}, practical sampling challenges with small patients \citep{whiteheadInterventionsPreventIatrogenic2019}, as well as smaller cohort sizes which limits the applicability of adult-derived models~\citep{shahUseMachineLearning2023}. While ML has been applied to related pediatric tasks such as infection detection \citep{martin2022machine,lee2022development,fenta2024factors, clarke2022future}, antibiotic susceptibility prediction \citep{oonsivilaiUsingMachineLearning2018}, and clinical deterioration \citep{velezEarlyPredictionAntibiotic2025}, to our knowledge no prior work has studied AMS intervention prediction models in the PICU.
\newline

\noindent To address these gaps, we provide a systematic, cross-cohort evaluation of AMS intervention prediction in PICU, using both a public and a private dataset. We jointly examine target formulation, temporal representation, and modeling paradigm, enabling a direct comparison of tabular, sequence, and graph-based models across clinically relevant stewardship tasks.


\section{Methods}
\label{sec:methods}

\begin{figure}[h]
  \centering
  \includegraphics[width=\linewidth]{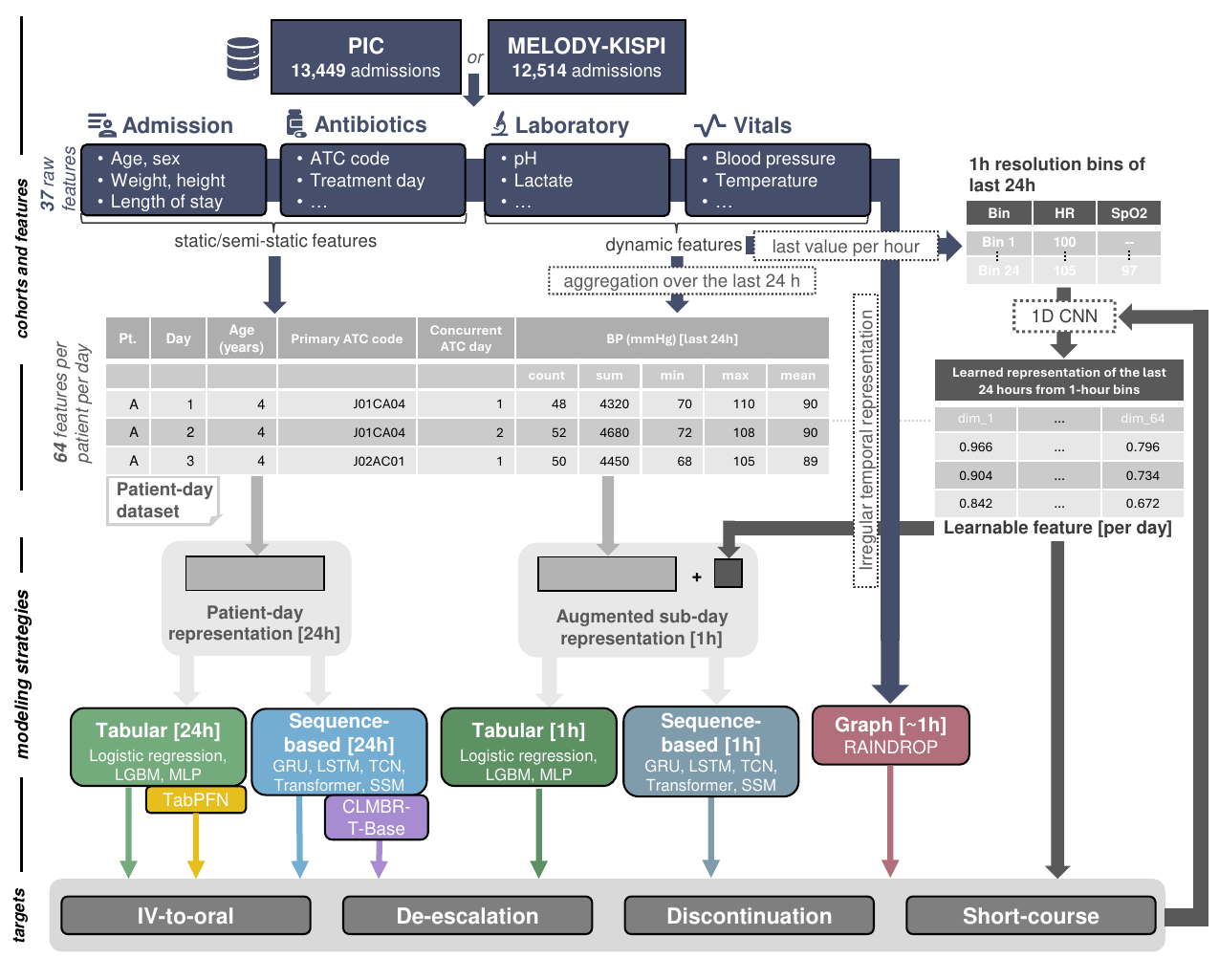}
  \caption{Benchmarking overview. EHR are represented at two temporal resolutions: (i) A X-day lookback tabular representation based on summary statistics of dynamic variables over the preceding day(s), and (ii) hourly bins  augmented by a 1-hour representation capturing intra-day dynamics through a learned embedding of hourly bins using a CNN trained jointly with the prediction task. We benchmark three classes of models under a unified framework. All models are evaluated on the four antimicrobial stewardship intervention targets.}
  \label{fig:overview}
\end{figure}

We formulate AMS intervention prediction as a patient-day classification task using EHR data. At the start of each antibiotic patient-day we want to predict whether an AMS intervention is likely to occur during that day. This section describes the prediction setting, data representation pipeline, model classes, and evaluation protocol (also see Figure~\ref{fig:overview}), whereas cohort-specific preprocessing details are provided in Section~\ref{sec:cohorts}. The code is made publicly available for reproducibility.\footnote{ \url{https://github.com/pinnacle-kispi/AMS_intervention_prediction}
}

\subsection{Prediction setting}

Predictions are made once a day at a fixed time, using only information available up to that point. Each prediction is defined for a patient-day with active antibiotic therapy and aims to predict whether an intervention occurs within the following 24 hours.
For tabular models, predictions are based on a single patient-day feature vector summarizing the most recent patient state. Temporal models use the same patient-day representation at each time step, but operate on sequences of consecutive patient-days, allowing them to incorporate information from prior days. This unified formulation ensures that all models receive the same per-day input, while differing only in their ability to leverage longitudinal context.


\subsection{Feature space}

We extract a consistent set of clinical variables across cohorts, including (i) patient and admission context, such as age and admission type, (ii) physiological measurements, namely vital signs and laboratory values, and
(iii) antibiotic exposure features (e.g., ATC code, route, concurrent therapies).
Full listing of all input features is available in Appendix Table~\ref{tab:extracted_data} alongside more detailed descriptions.

\subsection{Patient-day and temporal representations}
\label{patient-day and temporal representations}
We formulate the task at the patient-day level, where each instance corresponds to a 24-hour interval ending at a fixed daily prediction time. This defines a consistent prediction unit across all models.
We consider multiple temporal resolutions of this patient-day representation to disentangle the effect of temporal granularity and modeling assumptions (see Figure~\ref{fig:overview}). These representations provide different levels of temporal detail over the same prediction unit, enabling a controlled comparison of modeling approaches.

\paragraph{Patient-day representation (24h resolution).}
Physiological measurements are aggregated into summary statistics over the preceding 24 hours (e.g., minimum, maximum, mean) and combined with static and semi-static features, yielding a single feature vector per patient-day of the recent state. This corresponds to the standard tabular representation.

\paragraph{Augmented sub-day representations (1h resolution).}
To capture intra-day dynamics, we construct hourly representations within the same 24-hour window. Observations are aggregated into hourly bins by retaining the last recorded value in each bin. Empty bins are then forward-filled using last observation carried forward (LOCF) \citep{engelsImputationMissingLongitudinal2003}. We then concatenate the hourly event grid with the aggregated patient-day representation. A convolutional encoder summarizes this sequence into a compact embedding, which is concatenated with the 24h-resolution feature vector. As a result, the 1h representation augments the 24h representation, allowing models to jointly leverage coarse summary statistics and fine-grained temporal structure. The embedding is thus learned jointly with the prediction task. This representation allows tabular models to incorporate temporal information, and enables direct comparison across model classes.

\paragraph{Irregular temporal representation.}
We additionally consider a representation that preserves the native irregular sampling of EHR data within each patient-day. This is used by graph-based models, which directly operate on irregular observations and capture temporal and inter-variable relationships without discretization.

\subsection{Model classes}

We evaluate three classes of models, each corresponding to a different inductive bias about how clinical decisions arise from patient data. Implementations details can be found in Appendix~\ref{app:methods_impl_details}.

\paragraph{Tabular models.}
Tabular models operate on independent patient-day feature vectors and assume that relevant information is captured by the current state (i.e., previous patient-day). In line with previous work on AMS intervention prediction \citep{goodman2022paf, bystritsky2020asp, tran2024scaling, bolton2024personalising}, we include logistic regression \citep{mccullaghGeneralizedLinearModels2019}, LightGBM \citep{ke2017lightgbm}, and a multi-layer perceptron (MLP) classifier. We also include the transformer-based tabular foundation model TabPFN \citep{hollmann2025tabpfn} as a strong  baseline.

\paragraph{Sequence-based temporal models.}
Sequence models explicitly capture temporal dependencies across patient trajectories. They process patient data as sequences of observations and incorporate information from previous timesteps (here, patient-days) to inform predictions; either through recurrent state updates or through architectures that aggregate context across the sequence. 
We evaluate recurrent architectures (GRU, \citep{cho2014learning}, LSTM \citep{hochreiter1997lstm}), a Temporal Convolutional Network (TCN) \citep{bai2018empirical}, a Transformer
\citep{vaswani2017attention}, and a State Space Model (SSM) inspired by Mamba \citep{guMambaLinearTimeSequence2024}.
We additionally evaluate CLMBR-T-Base \citep{wornow2023ehrshot, guo2024multi}, a transformer foundation model pretrained on structured adult EHR code sequences. 

\paragraph{Graph-based temporal models.}
Graph-based temporal models provide an alternative to tabular and sequence-based approaches by representing clinical variables and observations as nodes, modeling their dependencies through message passing. Some architectures operate directly on irregular and asynchronous measurements, without fixed-interval discretization. In this work, we include RAINDROP \citep{Raindrop}, a state-of-the-art graph neural network designed for irregular multivariate time series classification. While newer classification methods, such as Hi-Patch \citep{luoHiPatchHierarchicalPatch2025} or WaveGNN \citep{hajisafiWaveGNNIntegratingGraph2025} have been proposed, they are not yet established as standard baselines. RAINDROP captures temporal and inter-variable dependencies through neural message passing and self-attention mechanisms.

\subsection{Targets}

\begin{figure}[h]
  \centering
  \includegraphics[width=0.7\linewidth]{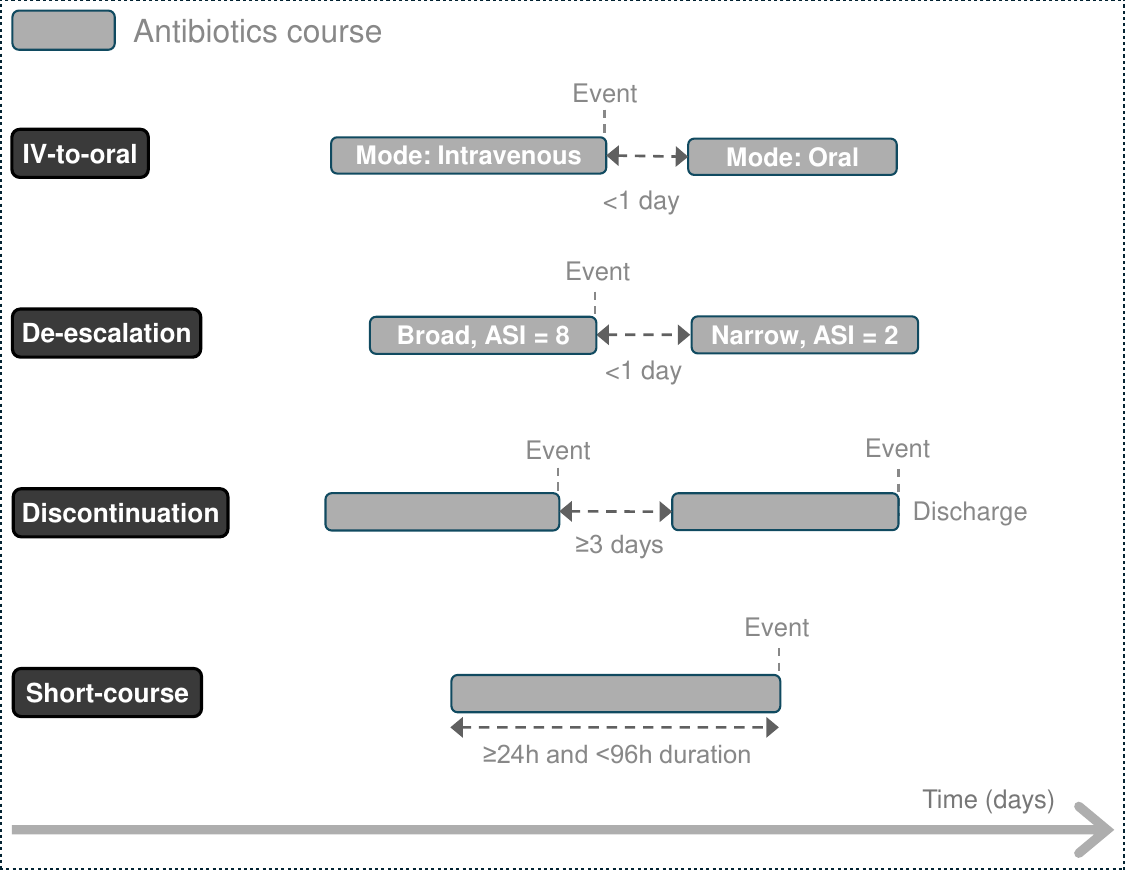}
  \caption{AMS intervention targets.}
  \label{fig:ams_targets}
\end{figure}

We define four AMS proxy targets derived from antibiotic treatment trajectories (see Figure \ref{fig:ams_targets} below and Appendix Section \ref{app:targets} for details):
\begin{itemize}
    \item \textbf{IV-to-oral}: Transition from intravenous to non-intravenous formulation within a treatment course.
    \item \textbf{De-escalation}: Switch to an antibiotic with a strictly lower antibiotic spectrum index (ASI) \citep{gerber2017asi}.
    \item \textbf{Discontinuation}: Termination of antibiotic therapy with at least 72 hours remaining before hospital discharge \citep{tran2024scaling}.
    \item \textbf{Short-course}: Total antibiotic duration longer than 24 hours and shorter than 96 hours, excluding prophylactic cefazolin.
\end{itemize}
Each target is defined at the patient-day level, and labeled positive only on the day the event occurs. IV-to-oral and de-escalation correspond to established stewardship actions aimed at reducing treatment invasiveness and narrowing the antimicrobial spectrum. Discontinuation and short-course targets capture opportunities to limit overall antibiotic exposure.

\paragraph{Target design.}
We build on prior work that operationalizes stewardship interventions from retrospective prescription data~\citep{tran2024scaling}, but adapt these definitions to the PICU setting and extend them in two important ways.
First, we introduce \textit{short-course} therapy as an additional proxy outcome. In pediatric critical care, treatments are often initiated per default at admission and then de-escalated early \citep{chiotos2022picu, abbas2016picu}. Short courses may therefore reflect situations where antibiotic therapy was initiated under uncertainty and subsequently discontinued when deemed unnecessary, providing a complementary signal for potential over-treatment.
Second, we define de-escalation strictly in terms of a reduction in ASI rather than also a reduction in the number of agents. This distinction is important: Reducing the number of antibiotics does not necessarily correspond to narrower antimicrobial coverage \citep{ilgesSpectrumScoresBetter2023}. By relying on spectrum-based definitions, we better align the target with stewardship intent and avoid conflating treatment simplification with true de-escalation.

\paragraph{Interpretation as intervention opportunities.}
These targets correspond to observed stewardship interventions and reflect real-world clinical decision-making, rather than adjudicated stewardship intent or optimal treatment decisions. Because they are derived from retrospective clinical practice, they capture when clinicians chose to modify or discontinue therapy-not necessarily when such actions were clinically appropriate. Consequently, models trained on these targets are likely to learn local prescribing and stewardship patterns, including potential biases, and may preferentially identify situations that resemble past intervention decisions. The direction and magnitude of this bias is difficult to quantify and may vary across institutions, clinical teams, and stewardship programs.

Despite this limitation, such proxies remain valuable for a CDSS. Rather than identifying definitive opportunities for antimicrobial optimization, the models are intended to flag patient-days that may warrant stewardship review. This approach provides a scalable and reproducible way to identify opportunities for prospective audit and feedback workflows, enabling more frequent and targeted assessment of antimicrobial therapy. 

\subsection{Evaluation}

Datasets are split per patient-day into training, validation, and test sets (70-10-20), ensuring that all stays of a given patient are assigned to the same split. The test-set is split at the beginning and not used until final evaluation. Models are trained using the same evaluation protocol across all model classes to ensure fair comparison.
Performance is evaluated at the patient-day level using standard classification metrics, including Area Under the Receiver Operating Characteristic (AUROC) \citep{hanley1982meaning}, F1 Score \citep{vanrijsbergen1979information}, and Area Under the Precision Recall Curve (AUPRC) \citep{davis2006relationship}.  

\section{Cohorts}
\label{sec:cohorts}

\subsection{Cohort Selection}
To evaluate AMS intervention prediction across heterogeneous pediatric settings, we use two complementary PICU cohorts: A publicly available dataset and the private \emph{MELODY-KISPI} dataset. This dual-cohort design enables both reproducibility and assessment of cross-site comparison, addressing a key limitation of prior work which typically relies on a single type of data source.
In both cohorts, we extract antibiotic treatment trajectories and construct patient-day observations corresponding to days on which antibiotic therapy is active, following the prediction setting described in Section~\ref{sec:methods}.

\paragraph{PIC.}
The Chinese Pediatric Intensive Care (PIC) database \citep{PicDB} is a publicly available dataset containing de-identified EHR data from patients admitted to the PICU of the Children’s Hospital of Zhejiang University School of Medicine between 2010 and 2018. It includes demographics, vital signs, laboratory measurements, and medication records, with physiological variables primarily recorded manually.  
We include all patients with at least one antibiotic prescription during their stay, resulting in 5,671 admissions for 5,515 distinct patients. The cohort is characterized by a young population (median age 0.83 years, IQR 0.16-3.81) and relatively long ICU stays (median 10.16 days, IQR 4.59-19.52). As a public dataset, PIC provides a reproducible benchmark for method comparison.

\paragraph{MELODY-KISPI.}
The \emph{MELODY-KISPI} cohort comprises de-identified EHR data from PICU admissions at the University Children's Hospital Zurich, Switzerland between 2015 and 2023. The retrospective use of these data was approved by the Cantonal Ethics Committee of Zurich\footnote{BASEC 2023-02077}. In contrast to PIC, this dataset reflects a different clinical environment and data collection process, including both sensor-derived and nurse-validated physiological measurements.
After applying the same inclusion criteria, the cohort contains 2,789 admissions for 2,059 distinct patients, with a younger and shorter-stay population (median age 0.40 years, IQR 0.04-2.91; median length of stay 5.18 days, IQR 1.77-15.78). This cohort enables evaluation of model robustness and generalization across institutions.

\subsection{Cohort-specific characteristics of AMS targets}

The distribution of AMS intervention targets differs substantially across cohorts and highlights important pediatric-specific considerations. In particular, intravenous-to-oral switching, commonly used as a proxy for stewardship interventions in adult settings, is extremely rare in both PICU cohorts (0.34\% in PIC and 0.98\% in the \emph{MELODY-KISPI} cohort, see Table \ref{tab:cohort_summary} in the Appendix). 
In contrast, short-course therapy is substantially more frequent, especially in the \emph{MELODY-KISPI} cohort, and often co-occurs with discontinuation, suggesting that it captures a meaningful and prevalent pattern of early treatment cessation. Other combinations of intervention types are rare, indicating that stewardship actions are typically isolated events (see  Figure~\ref{fig:UpSet_overlap_outcomes} in the Appendix).
These differences motivate the inclusion of short-course therapy as an additional proxy outcome tailored to the pediatric setting, complementing established intervention targets and enabling a broader characterization of opportunities for reducing unnecessary antibiotic exposure.

\subsection{Data Extraction}
We extract a set of 37 clinical variables from both cohorts, covering demographics, admission context, antibiotic prescriptions, vital signs, and laboratory measurements (see Table~\ref{app:data} in the Appendix for the full list). 
To enable consistent modeling and fair comparison, both datasets are mapped to a shared feature space covering demographics, clinical measurements, and antibiotic exposure. Antibiotics records are limited to an overlapping set and harmonized using ATC \citep{who_atc_2024} codes (see Table~\ref{tab:supp_antibiotic_spectrum} in the Appendix). We assume that prescription timestamps approximate administered therapy in PIC. For the \emph{MELODY-KISPI} cohort, the available timestamps correspond to actual antibiotic administrations. Laboratory measurements are aligned using LOINC codes \citep{forreyLogicalObservationIdentifier1996}.

\subsection{Feature construction}

We apply the unified feature construction pipeline described in Section~\ref{sec:methods} to both cohorts, ensuring that differences in performance reflect underlying data characteristics and not preprocessing choices. Here, we highlight cohort-specific implementation details. Antibiotic prescriptions are mapped to systemic antibiotics and merged into clinically contiguous courses based on temporal proximity (within 24 hours). This step is particularly important for the \emph{MELODY-KISPI} cohort, where prescriptions were mostly recorded as individual bolus administrations as opposed to continuous treatments. Very short courses ($<24$ hours), typically reflecting prophylactic use, are excluded. When multiple antibiotic courses overlap within a patient-day, all active agents are retained, jointly contributing to feature construction and target labeling, instead of collapsing to a single representative treatment.
For the \emph{MELODY-KISPI} cohort, antibiotic prescription timestamps correspond to actual administrations, while in PIC they approximate administered therapy. This difference does not affect feature construction but is relevant for interpreting temporal alignment.
All remaining preprocessing and feature aggregation steps follow the shared pipeline described in Section~\ref{sec:methods}.

\section{Experiments and Results}
\label{sec:results}


\subsection{Comparison to previous work in adults}

For contextualization, we compare the results of our tabular models trained on 24-hour aggregated pediatric data (LightGBM and MLP) with previously reported adult ICU results (full results in Table~\ref{tab:tran_the_bolton_comparison} in Appendix \ref{app:comparison_to_adults}). The feature set used is available to both pediatric datasets (Appendix Section~\ref{app:data}), comparable to \citet{tran2024scaling} and includes 7/10 of the base vital features used in  \citet{bolton2024personalising}. 
Intervention prevalence differs substantially between adult and pediatric settings. IV-to-oral switching is much rarer in PICU cohorts (0.3-1.0\%) than in adult data (3.0\%), and discontinuation is also less frequent (1.7-5.7\% vs 9\%). We observe high AUROC values in pediatric cohorts (e.g., 0.93 for de-escalation), but precision-recall performance is markedly lower for rare targets. For IV-to-oral for example, LightGBM shows low AUPRC and TPR scores compared to \citet{tran2024scaling}, reflecting extreme class imbalance. This highlights that AUROC alone can overestimate performance in such settings. For more prevalent targets, strong AUROC does not consistently translate to improved AUPRC, indicating differences in label distribution and clinical practice. Overall, these results show that performance does not directly transfer from adult to pediatric cohorts. Model behavior is strongly influenced by target prevalence and cohort characteristics, motivating the need for PICU-specific evaluation and the inclusion of additional targets such as short-course therapy.

\subsection{Effect of sequence modeling at 24-hour temporal resolution}

We then investigate whether sequence-based models can better exploit longitudinal structure in patient trajectories than tabular models when both use the same 24-hour aggregated input representation. Specifically, we compare sequence-based models (recurrent, convolutional, and attention-based) with tabular baselines, using the same 24-hour aggregated representation across the \emph{PIC} and \emph{MELODY-KISPI} cohorts. AUROC, F1 score and AUPRC are shown in Figure~\ref{fig:comparison_pic_private_24h}, with full results provided in Appendix~\ref{app:detailed_results_targets}. We report the performance of the pretrained models TabPFN and CLMBR-T-Base separately because, unlike the other models, they are not trained from scratch on these cohorts. The F1 score threshold is optimized using the validation split, and averages of three seeds are always reported on the held-out test set in the results below.

\begin{figure}[h]
    \centering
    \includegraphics[width=1\linewidth]{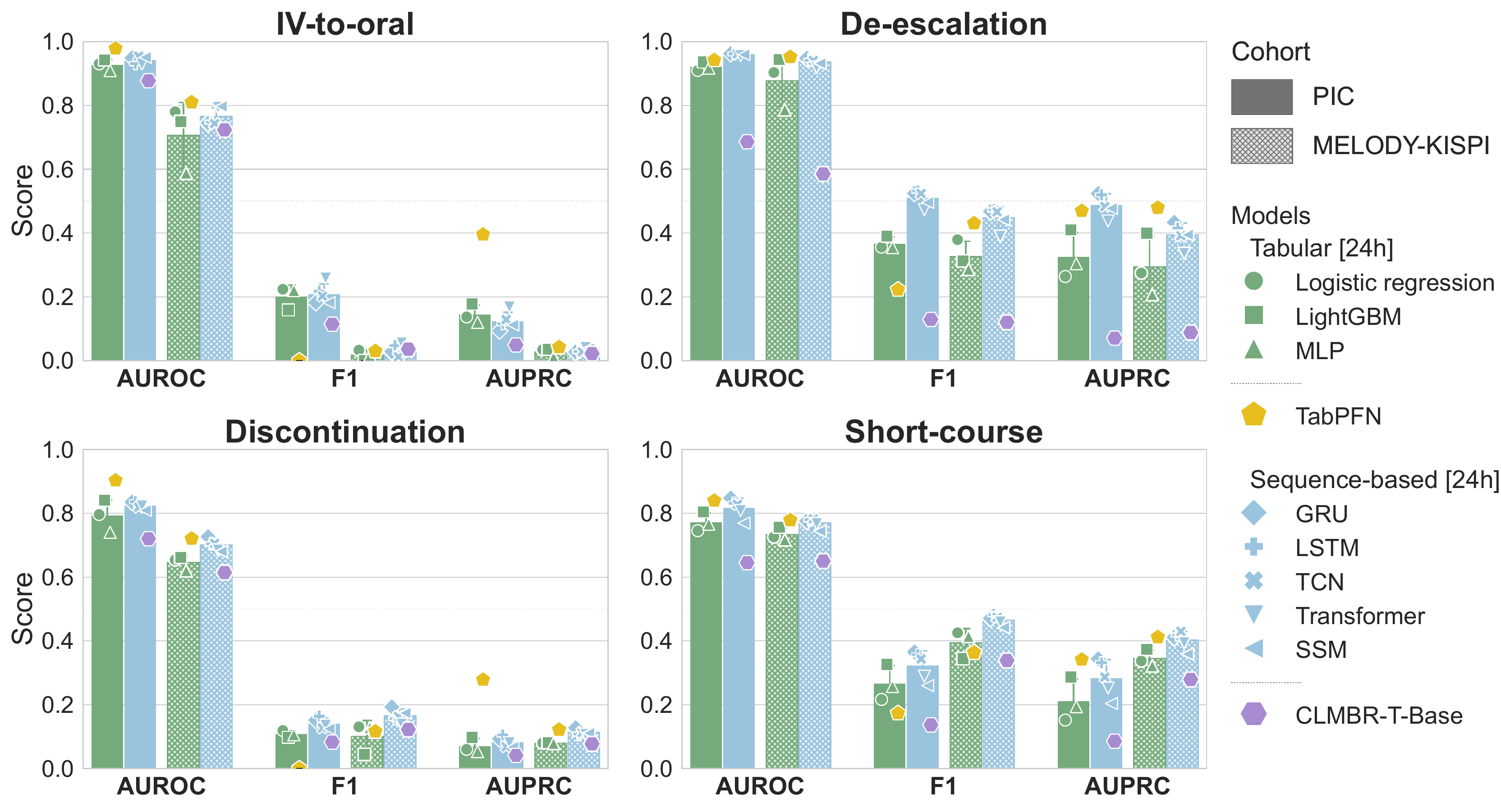}
    \caption{Comparison of tabular and sequence models using 24-hour aggregated representations across the \emph{PIC} and \emph{MELODY-KISPI} cohorts and across four antimicrobial stewardship targets. Bars show the mean performance within each model group. Individual model results are overlaid as points. The pretrained models TabPFN and CLMBR-T-Base are shown separately for reference.}
    \label{fig:comparison_pic_private_24h}
\end{figure}
Across tasks and cohorts, sequence models achieve similar AUROC to tabular models (e.g., $\sim$0.85-0.95 for IV-to-oral and de-escalation), indicating that most predictive signal is already captured by the current patient-day representation.
Differences are more apparent in AUPRC and threshold-optimized F1 score. For rare targets (IV-to-oral, discontinuation), sequence models provide modest gains over tabular models, while for more prevalent targets (de-escalation, short-course), the gains are substantially larger. Similar trends are observed across both cohorts. In \emph{PIC}, the average F1 score increased from 0.33 across tabular models to 0.51 across sequence models for de-escalation, and from 0.24 to 0.32 for short-course. Smaller but consistent gains are also observed for \emph{MELODY-KISPI}. F1 score differences were even larger at the default threshold of 0.5 (Appendix~\ref{app:detailed_results_targets}). The results suggest that explicit temporal modeling can improve the prediction of interventions over tabular approaches, and that exploring finer-grained temporal representations may offer additional improvements.

\subsection{Effect of finer temporal resolution and irregular time series modeling}

\begin{figure}[h]
    \centering
    \includegraphics[width=0.75\linewidth]{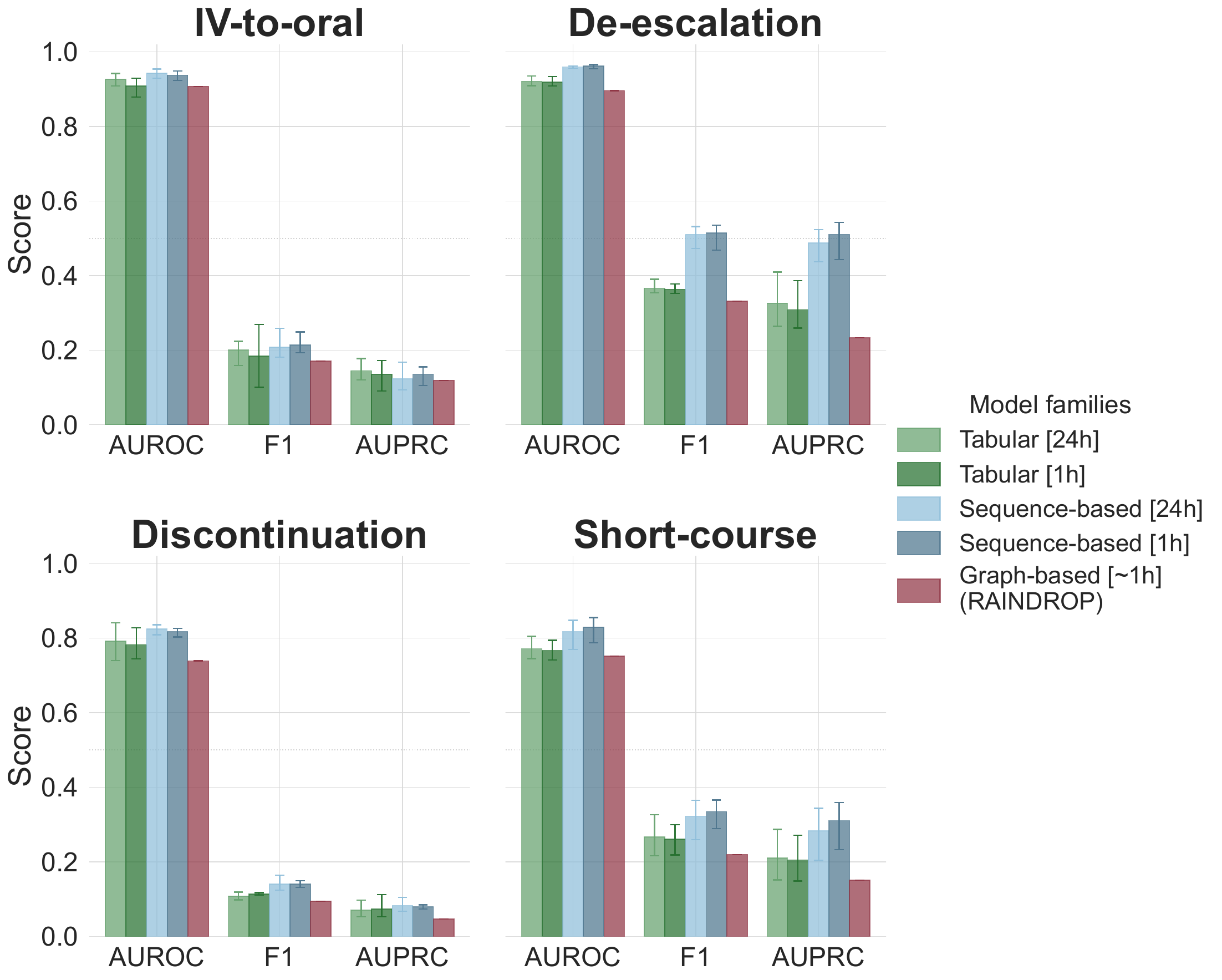}
    \caption{Performance of tabular and sequence-based models at 24-hour resolution and with augmented 1-hour representation, together with the graph-based model RAINDROP on irregular time series on the \emph{PIC} cohort. Bars and error bars show the model class mean and minimum-maximum range, respectively.}
    \label{fig:comparison_24_1h_pic}
\end{figure}

We next evaluate whether increasing temporal resolution improves performance. We augment 24-hour features with 1-hour event grids (See Section~\ref{patient-day and temporal representations}) and compare tabular and sequence models, together with the graph-based model (RAINDROP) on the \textit{PIC} cohort (Figure~\ref{fig:comparison_24_1h_pic}). We exclude the pretrained TabPFN and CLMBR-T-Base models from this follow-up experiment because they performed poorly in the preceding 24-hour-resolution comparison. Corresponding results for the \emph{MELODY-KISPI} cohort are available in Figure~\ref{fig:Comparison_24h_1h_private} of the Appendix.
Incorporating finer-grained temporal information only slightly improves performance. For de-escalation and short-course, F1 and AUPRC for de-escalation and short-course targets improve $\sim$0.01-0.03 on average across three seeds in \emph{PIC}, and similar results were observed in \emph{MELODY-KISPI} (Appendix Figure \ref{app:calibration_Private}). The CNN-based learnable encoder does add some predictive power over a simple linear projection or direct hourly modeling (Appendix Section \ref{app:1h ablations}, Figures \ref{fig:1h_ablations_pic} and \ref{fig:1h_ablations_private}). Tabular models augmented with temporal embeddings show less consistent results and remain below sequence-based approaches. RAINDROP, also modeling independent patient-days, shows performance comparable to but below that of the tabular models, suggesting that the graph-based embeddings of irregularly sampled data confuse the predictors for these tasks.

Overall, these results indicate that temporal resolution and structure matter, although they provide only limited incremental benefits, and fully leveraging them requires models that can naturally handle sequences as opposed to static snapshots.

\subsection{Evaluating reliability across model classes}
Trust is a critical factor in the adoption of a CDSS for AMS \citep{bolton2025impact}. In this context, well-calibrated predictions are essential, as they determine how reliably predicted probabilities reflect true intervention likelihood. We therefore evaluate calibration across tabular, sequence-based, and graph-based temporal models using different temporal representations. Results for the \emph{PIC} cohort are shown in Figure~\ref{fig:calibration_PIC}, with corresponding results for \emph{MELODY-KISPI} provided in Appendix Figure~\ref{fig:calibration_private}.
\begin{figure}[h]
    \centering
    \includegraphics[width=1\linewidth]{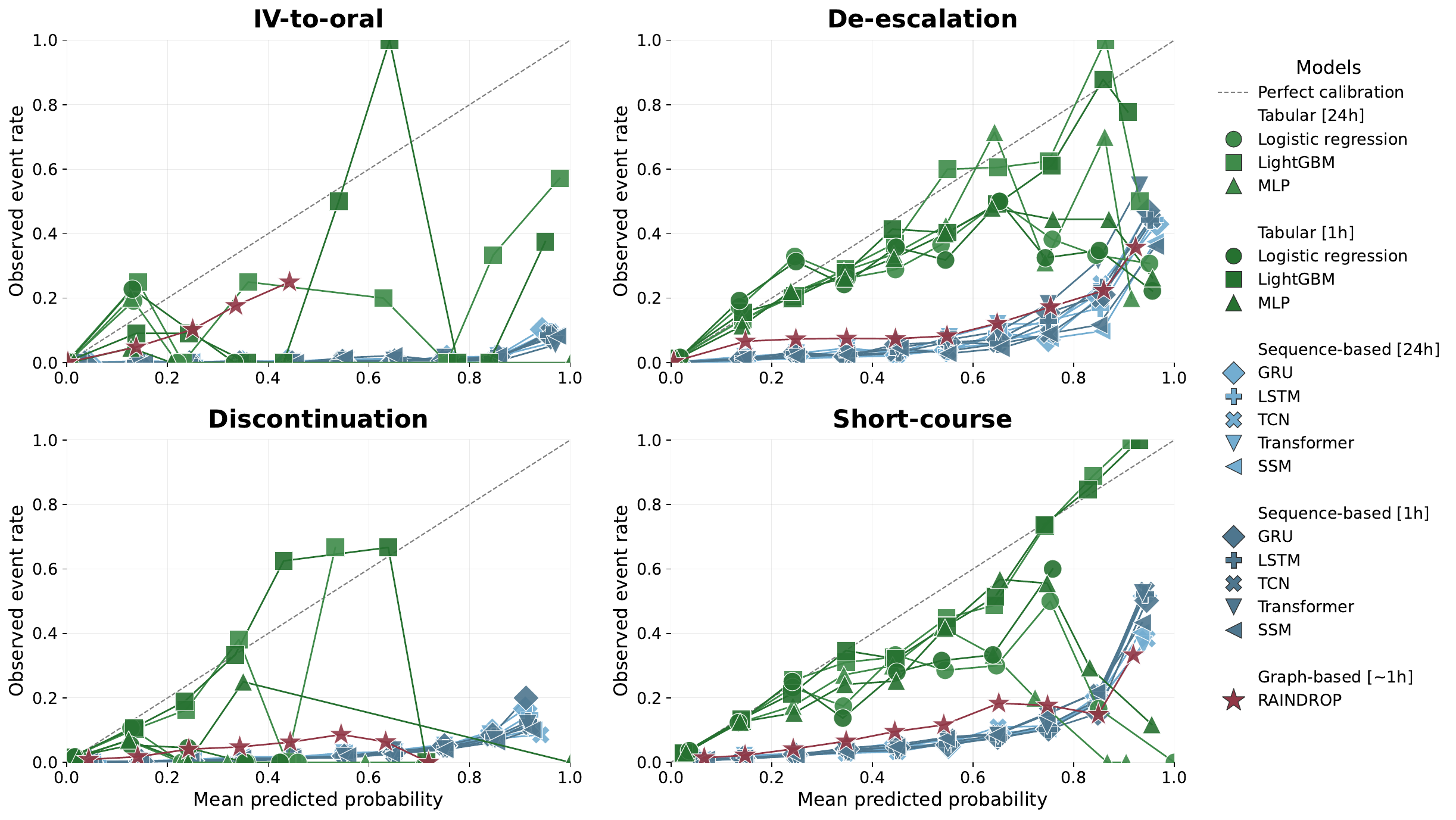}
    \caption{Calibration plots for tabular, sequence-based, and graph-based models across temporal representations for the four AMS intervention predictions tasks in the \emph{PIC} cohort.}
    \label{fig:calibration_PIC}
\end{figure}
Calibration varies substantially across model classes and tasks. Tabular models, particularly LightGBM, show the best calibration, with predicted probabilities closely aligned with observed event rates for more prevalent targets, such as de-escalation and short-course. In contrast, sequence models are over-confident, with predicted probabilities consistently higher than the observed event rates across bins. Complementary experiments highlight that predicted probabilities generally increase more towards the event day for sequence models compared to tabular ones (Figures \ref{fig:event_aligned_probabilities_pic} and \ref{fig:event_aligned_probabilities_private} in the Appendix). 
We note that RAINDROP exhibits similar calibration challenges, with weak probability separation and poor calibration. Importantly, increasing temporal resolution does not substantially improve calibration. Models using 1-hour representations exhibit similar over-confidence as their 24-hour counterparts, indicating that calibration is largely independent of temporal resolution in this setting. 
For rare targets (IV-to-oral, discontinuation), all models show unstable calibration, with noisy estimates and unreliable behavior at higher probability ranges due to extreme class imbalance. However, for more prevalent targets, like short-course in the \emph{MELODY-KISPI} cohort (Appendix Figure~\ref{fig:calibration_private}), we observe that the sequence models and RAINDROP recover calibration. 
Overall, these results highlight a trade-off between discrimination and calibration. While more complex temporal models can improve ranking performance, simpler tabular models provide more reliable probability estimates, which is essential for building a trustworthy CDSS.

\subsection{Additional Results}
We use patient-random splitting for consistency, as \emph{PIC} employs random time offsets that preclude patient-temporal splitting. On \emph{MELODY-KISPI}, we compared both strategies on held-out test sets and found that patient-temporal splitting consistently yields lower performance, suggesting that patient-random splitting provides more optimistic estimates (see Table~\ref{tab:split_strategy_private} in the Appendix).
Furthermore, we assume patient-days are independent which may inflate the effective sample size due to within-admission correlation. However, nearly identical na\"ive and admission-level cluster bootstrap confidence intervals (Appendix Table \ref{tab:within_admission_correlation_performance_metrics}) indicate that this correlation has negligible impact on performance uncertainty. 
Because of conceptual overlap between AMS actions and prior modeling efforts that did not distinguish intervention targets \citep{bystritsky2020asp, goodman2022paf}, we also explored multi-task learning in Appendix Section \ref{app:mtl}. However, it provided limited benefit, likely due to limited stay-level target overlap (Appendix Figure \ref{fig:UpSet_overlap_outcomes}), hierarchical relationships between targets, and differences in label quality. These findings support task-specific modeling, particularly for common or heterogeneous intervention targets.

\section{Discussion}
In this work, we systematically evaluate multiple modeling paradigms for AMS intervention prediction in the PICU across cohorts, targets, and temporal representations, yielding several key insights.

First, we observed that performance was driven by data characteristics and sequential memory rather than model complexity. Outcome prevalence and temporal structure largely determined both ranking and absolute performance.
In particular, IV-to-oral antimicrobial switching was uncommon in our PICU cohorts, consistent with the clinical realities of pediatric critical care \citep{berthe2012antibiotic, mcmullan2016ivoral_systematic}, highlighting the limited transferability of models developed in adult cohorts. 
%
One reason for the low prevalence of this target in pediatric data, could be the limited number of oral administrations in PICU. Intravenous administration accounted for 85-98\% of antimicrobial courses, likely reflecting sedation, mechanical ventilation, gastrointestinal dysfunction, limited antimicrobial formulations available in both intravenous and oral forms, and the very young age of the cohorts (median age: \emph{PIC} $\sim$10 months; \emph{MELODY-KISPI} $\sim$3 months). This was also reflected in treatment initiation, where patients receiving intravenous therapy were younger than those receiving oral therapy in both cohorts: in \emph{PIC}, median age at course start was 10.8 months (IQR 2.0–48.4) for IV versus 14.6 months (IQR 2.2–60.4) for oral therapy; in the \emph{MELODY-KISPI} cohort, median age was 5.0 months (IQR 0.5–42.3) for IV versus 8.7 months (IQR 0.1–26.8) for oral therapy.
Despite its low frequency in the PICUs, IV-to-oral switching remains an important AMS intervention strategy and should be routinely considered during treatment review in critically ill patients \citep{willemsOptimizingUseAntibiotic2021}. Oral therapy reduces catheter-related infections and thrombosis, pain and anxiety associated with intravenous access, hospital length of stay, and healthcare costs \citep{cotterOpportunitiesStewardshipTransition2021,kerenComparativeEffectivenessIntravenous2015,shahIntravenousOralAntibiotics2016,christensenEffectsHospitalPractice2019,ruebnerComplicationsCentralVenous2006,girdwoodImprovingTransitionIntravenous2020}. Although switching is often limited by clinical instability, impaired gastrointestinal absorption, and restricted oral treatment options, it is feasible in carefully selected children who are clinically improving, able to tolerate enteral therapy, and have an appropriate oral antimicrobial available \citep{mcmullan2016ivoral_systematic,willemsOptimizingUseAntibiotic2021}.
The low prevalence of IV-to-oral switching likely contributed to the relatively low F1 scores and AUPRC observed across models despite reasonable AUROC, underlining the challenges of predicting rare clinical events. Nevertheless, the similar data characteristics and model performance observed across both PICU cohorts suggest that development of generalizable pediatric-specific AMS prediction models is feasible.

Second, we noticed that more fine-grained temporal representations provided only modest additional gains beyond 24-hour aggregated features, suggesting that much of the predictive signal was already captured by daily summaries. This supports the use of simpler time-agnostic model development, such as done by \citet{tran2024scaling} and \citet{bolton2024personalising}. Nevertheless, sequence models consistently improved F1 score and AUPRC over tabular approaches, indicating that longitudinal context remains useful for intervention prediction. RAINDROP did not outperform the tabular or sequence-based model classes, providing no clear evidence that graph-based temporal modeling of irregularly sampled events benefits these tasks. Its underperformance may partly reflect our computationally motivated design choice to train the models on independent patient-days, thereby limiting temporal context. However, because RAINDROP also underperformed tabular models with the same patient-day context, limited context alone is unlikely to fully explain the result. Further work is needed to determine whether the difference arises from model complexity, optimization, or data scale. Moreover, the limited performance of CLMBR-T-Base further suggests that representations pretrained on adult data may not transfer directly to pediatric cohorts, consistent with prior evidence of challenges in transferring adult-derived models to children~\citep{sabharwalCombiningAdultPediatric2022a}. 
Overall, the results highlight the importance of temporal representation alongside model architecture and pretraining population.

Lastly, we observe a trade-off between discrimination and reliability. While sequence-based models improve ranking performance, they are consistently overconfident compared to tabular models. Since trust and reliability are critical for the adoption of an antibiotic CDSS  \citep{lakaFactorsThatImpact2021, bolton2025impact}, and finer temporal resolution does not improve calibration, explicit post-hoc calibration may be necessary before deployment.

Overall, these findings suggest that, in the context of PICU AMS intervention prediction, target definition, temporal representation, and calibration are as critical as model choice, and increasing model complexity alone is insufficient for reliable clinical decision support.

\paragraph{Practical deployment considerations.}
We envision integrating AMS modeling into PICU practice through a dashboard providing daily risk predictions to support prioritization of ongoing antibiotic courses for stewardship review. Since AMS rounds at the University Children's Hospital Zurich are conducted twice weekly, these predictions could prompt ad hoc reviews and support earlier, more systematic stewardship interventions. LightGBM provided the best overall balance of discrimination, calibration, and interpretability. Calibrated GRU models may be preferred when intervention targets are common, identifying patients for review is particularly important, and antibiotic courses are longer (Appendix \ref{app:when_sequence_models_help}).

\paragraph{Limitations and future work.}
Importantly, the prediction targets are derived from retrospective stewardship decisions and therefore represent clinically enacted intervention opportunities. 
Although these events likely reflect AMS decisions, some may have occurred for other reasons than reducing unnecessary exposure, such as adverse effects or changes in diagnosis. The labels therefore represent proxy intervention events and not confirmed stewardship actions and may also fail to capture missed or earlier opportunities for intervention. A key next step is to assess their clinical validity through review by pediatric infectious disease specialists and PICU clinicians, using the full clinical record and notes to determine the indication, appropriateness, and timing of each observed change. Nonetheless,  \citet{tran2024scaling} already achieved better AMS review prioritization using the proxy labels.
After validation, a prospective evaluation similar to \citet{bolton2025impact} will be required to determine whether model-guided review improves stewardship efficiency, prescribing, and patient outcomes in practice.
We further observed that class imbalance substantially affected both discrimination and calibration, limiting the clinical utility of the models. Future work should therefore investigate uncertainty-aware methods, including Bayesian approaches, ensemble Bayesian model averaging, and conformal prediction, to improve calibration and provide reliable confidence estimates for a CDSS \citep{ruheBayesianModellingPractice2024,koumantakisImprovingMachineLearning2026,kalitaSimulationBasedValidationFramework2026}. 
Finally, our framework relies on fixed temporal aggregation windows, which may obscure clinically relevant patterns happening at a finer-grained resolution, while missing data are handled primarily through imputation. Although a recent study on PICU data found that LOCF imputation performed competitively and often outperformed more complex methods \citep{digitaleMethodsAddressingMissingness2025}, it may be less suitable during periods of rapid physiological change \citep{lachinFallaciesLastObservation2016}, commonly encountered in the PICU. Future work should therefore explore missingness-aware temporal models, such as GRU-D \citep{cheRecurrentNeuralNetworks2018}, which explicitly model informative missingness and have shown promise for clinical time-series prediction\citep{liptonModelingMissingData2016}.

\section{Acknowledgments}
This research project and related results were made possible with the support of the NOMIS foundation. Niklas Raehse and Daphn\'e Chopard are both supported by the grant "Using artificial intelligence to improve early recognition and treatment of critically ill children (AI in pediatric ICU)" from the NOMIS foundation.
We would also like to thank Catherine Jutzeler from ETH Zurich for providing the supervisory and institutional infrastructure to enable this project and Philipp Baumann from the University Children's Hospital Zurich, for his support in contextualizing the study and providing valuable clinical insights.

\bibliography{references}

\newpage
\appendix

\section{Data} 
\label{app:data}

\subsection{Cohorts}
\label{app:cohorts}
\subsubsection{PIC}
PIC version 1.1.0 \citep{PicDB} was retrieved from Physionet \citep{PhysioNet}.

\subsubsection{MELODY-KISPI}
The PICU dataset from the University Children’s Hospital Zurich, referred to as MELODY-KISPI, includes patients younger than 18 years who were admitted between 1 January 2015 and 31 August 2023. The retrospective use of these data was approved by the Cantonal Ethics Committee of Zurich under the MELODY project ("Machine Learning in the Paediatric Intensive Care Unit: Development of Risk Prediction Models"; BASEC ID 2023-02077). In accordance with the ethics approval, patients with a documented refusal of general consent were not included.

\subsection{Extracted data}
\vspace{-0.4in}
\begingroup
\small
\setlength{\tabcolsep}{5pt}
\renewcommand{\arraystretch}{1.15}
\begin{longtable}{@{}
  >{\raggedright\arraybackslash}p{0.24\textwidth}
  >{\raggedright\arraybackslash}p{0.18\textwidth}
  >{\raggedright\arraybackslash}p{0.25\textwidth}
  >{\raggedright\arraybackslash}p{0.25\textwidth}@{}}
\phantomsection\\
\caption{List of variables extracted from raw datasets and aggregated features included in the patient-day dataset.}\label{tab:extracted_data}\\

\toprule
\textbf{Raw variable} & \textbf{Source Table} & \textbf{Aggregated features} & \textbf{Notes} \\
\midrule
\endfirsthead
\multicolumn{4}{@{}l}{\textit{\tablename\ \thetable{} -- continued}}\\
\toprule
\textbf{Raw variable} & \textbf{Source Table} & \textbf{Aggregated features} & \textbf{Notes} \\
\midrule
\endhead
\midrule
\multicolumn{4}{r@{}}{\textit{Continued on next page}}\\
\endfoot
\bottomrule
\endlastfoot

\multicolumn{4}{@{}l}{\textbf{Patient and admission context}}\\
\midrule
Age at admission (years) & PATIENTS & & \\
Gender & PATIENTS & & One-Hot Encoding (OHE) \\
Ethnicity & ADMISSIONS & & OHE \\
Admission type & ADMISSIONS & & OHE \\
Insurance & ADMISSIONS & & OHE \\
Current LOS hours at day start & ADMISSIONS & & Visit trajectory context \\
Cumulative ICU hours & ADMISSIONS & icu\_hours\_cum & Visit trajectory context \\
\midrule

\multicolumn{4}{@{}l}{\textbf{Physiological measurements}}\\
\multicolumn{4}{@{}l}{\textit{Vital signs}}\\
\midrule
Temperature & CHARTEVENTS & sum, mean, n, min, max & PIC ITEMID: 1001 \\
Heart Rate & CHARTEVENTS & sum, mean, n, min, max & PIC ITEMID: 1003 \\
Respiratory Rate & CHARTEVENTS & sum, mean, n, min, max & PIC ITEMID: 1004 \\
Oxygen Saturation & CHARTEVENTS & sum, mean, n, min, max & PIC ITEMID: 1006 \\
Diastolic Pressure & CHARTEVENTS & sum, mean, n, min, max & PIC ITEMID: 1015 \\
Systolic Pressure & CHARTEVENTS & sum, mean, n, min, max & PIC ITEMID: 1016 \\
Height & CHARTEVENTS & sum, mean, n, min, max & PIC ITEMID: 1013 \\
Weight & CHARTEVENTS & sum, mean, n, min, max & PIC ITEMID: 1014 \\
\midrule

\multicolumn{4}{@{}l}{\textit{Trend features}}\\
\midrule
Vital trend features & CHARTEVENTS & delta\_prev\_day\_max, delta\_abx\_start\_max, var\_3d & Built for TEMP, RR, HR, SBP, DBP, SPO2. Day-to-day differences, deviations from antibiotic start, 3-day variance. \\
\midrule

\multicolumn{4}{@{}l}{\textit{Laboratory values}}\\
\midrule
Albumin & LABEVENTS & sum, mean, n & LOINC: 1751-1 \\
Alanine Aminotransferase (ALT) & LABEVENTS & sum, mean, n & LOINC: 1742-6 \\
Lactate Dehydrogenase & LABEVENTS & sum, mean, n & LOINC: 2532-0 \\
Hemoglobin & LABEVENTS & sum, mean, n & LOINC: 718-7 \\
Red Blood Cells & LABEVENTS & sum, mean, n & LOINC: 789-8 \\
Lactate & LABEVENTS & sum, mean, n & LOINC: 2532-0 \\
Sodium Whole Blood & LABEVENTS & sum, mean, n & LOINC: 2947-0 \\
PCO2 & LABEVENTS & sum, mean, n & LOINC: 11557-6 \\
Ph & LABEVENTS & sum, mean, n & LOINC: 11558-4 \\
PO2 & LABEVENTS & sum, mean, n & LOINC: 11556-8 \\
Oxygen Saturation & LABEVENTS & sum, mean, n & LOINC: 20564-1 \\
White Blood Count (WBC) & LABEVENTS & sum, mean, n, recent\_90d & LOINC: 26464-8 \\
C-Reactive Protein (CRP) & LABEVENTS & sum, mean, n, recent\_90d & LOINC: 32264 \\
Procalcitonin & LABEVENTS & sum, mean, n, recent\_90d & LOINC: 33959-8 \\
\midrule

\multicolumn{4}{@{}l}{\textbf{Antibiotic exposure features}}\\
\midrule
Primary ATC code & PRESCRIPTIONS & & Longest running antibiotic course class \\
Antibiotic medication features & PRESCRIPTIONS & distinct\_atc\_ever, concurrent\_atc\_day, any\_oral\_rx\_last\_3d & Antibiotic exposure dynamics: classes, concurrent count, recent oral use. \\

\end{longtable}
\endgroup

\subsection{Description of data and features}
\paragraph{Patient and admission context.} 
Static and semi-static features include age at admission, sex, ethnicity, admission type, and insurance status (Table~\ref{tab:extracted_data}). These features capture baseline patient characteristics and care context.
These are encoded using one-hot representations where applicable. We also include visit-level context such as length of stay at the start of the day and cumulative ICU time to capture disease progression and care trajectory.

\paragraph{Physiological measurements.} 
Time-varying clinical variables include vital signs (e.g., temperature, heart rate, respiratory rate, blood pressure, and oxygen saturation) and laboratory measurements (e.g., C-reactive protein (CRP), white blood cell count (WBC), and lactate). These features capture the evolving physiological state of the patient.
For the 24h aggregated data, we compute summary statistics over the pre-day window, including sum, mean, count, and where appropriate, minimum and maximum. For sparsely recorded inflammatory markers (e.g. CRP, procalcitonin), we additionally include longer-term summaries (e.g., most recent value in last 90 days) to capture baseline trends. To incorporate short-term dynamics within the tabular representation, we also compute engineered trend features for key vital signs (e.g. temperature, heart rate, respiratory rate, blood pressure, oxygen saturation), including day-to-day differences, deviations from antibiotic start, and short-window variability (e.g. 3-day (lookback) variance).

\paragraph{Antibiotic exposure features.}
We include features describing antibiotic treatment patterns, including current and prior antibiotic classes, route of administration, and concurrent therapies. These features provide context for the AMS interventions. 
We derive features describing antibiotic usage patterns from prescription data, including current and historical antibiotic classes (ATC codes), number of concurrent antibiotics, prior exposure to high-priority antibiotics, and recent oral antibiotic use. These features aim to capture treatment dynamics relevant for stewardship decisions. To support the cross-cohort comparison and the definition of the de-escalation outcome, we map antibiotics to ATC codes, product names, and their ASI scores \citep{gerber2017asi} in Table~\ref{tab:supp_antibiotic_spectrum} below.

\subsection{Antibiotics mapping}
\vspace{-0.4in}
\begin{longtable}{
@{}
p{0.6in}
p{1.6in}
p{0.3in}
p{3.0in}
@{}
}
\phantomsection\\
\caption{Antibiotics tracked for creating patient-days dataset. Mappings to the Anatomical Therapeutic Chemical (ATC) standard were used for extraction from the hospital information system of the University Children's Hospital Zurich. A mapping to the Antibiotic Spectrum Index (ASI) score \citep{gerber2017asi} and the corresponding product names in the PRESCRIPTIONS table of \emph{PIC} is included.}\label{tab:supp_antibiotic_spectrum}\\

\toprule
ATC & Label & ASI & PIC product name \\
\midrule
\endfirsthead
\multicolumn{4}{c}{\tablename\ \thetable{} --- continued} \\
\toprule
ATC & Label & ASI & PIC product name \\
\midrule
\endhead
\midrule
\multicolumn{4}{r}{\textit{Continued on next page}} \\
\endfoot
\bottomrule
\endlastfoot
J02AX01 & 5-fluorocytosine &  & --- \\
J01FA09 & acetylspiramycin & 4 & Clarithromycin Granule for Oral Suspension; Clarithromycin Tablets \\
J01GB06 & amikacin & 5 & Amikacin Sulfate Injection \\
J01CA04 & amoxicillin & 2 & Amoxicillin Sodium and Clavulanate Potassium for Injection; Amoxicillin and Clavulanate Potassium Granules; Amoxicillin Sodium and Sulbactam Sodium for Injection; Amoxicillin and Clavulanate Potassium Tablets \\
J01CR02 & amoxicillin/clavulanic acid & 6 & --- \\
J02AA01 & amphotericin b &  & Amphotericin B Liposome for Injection \\
J01CA01 & ampicillin & 2 & Ampicillin Sodium and Sulbactam Sodium for Injection; Ampicillin Sodium for Injection \\
J01CR01 & ampicillin/sulbactam & 6 & --- \\
J01FA10 & azithromycin & 4 & Azithromycin for Injection; Azithromycin for Suspension; Azithromycin Lactobionate for Injection; Azithromycin Tablets \\
J01DF01 & aztreonam & 5 & Aztreonam for Injection \\
J01DC04 & cefaclor & 4 & Cefaclor for Suspension; Cefaclor Capsules \\
J01DB04 & cefazolin & 3 & --- \\
J01DE01 & cefepime & 6 & Cefepime Hydrochloride For Injection \\
J01DD62 & cefoperazone/sulbactam & 6 & Cefoperazone Sodium and Sulbactam Sodium for Injection \\
J01DD01 & cefotaxime & 5 & Cefotaxime Sodium for Injection \\
J01DC05 & cefotetan & 4 & --- \\
J01DC01 & cefoxitin & 4 & --- \\
 & cefoxitin screen &  & --- \\
J01DD02 & ceftazidime & 4 & Ceftazidime for Injection \\
J01DD04 & ceftriaxone & 5 & Ceftriaxone Sodium for Injection \\
J01DC02 & cefuroxime & 4 & Cefuroxime Sodium For Injection; Cefuroxime Sodium for Injection; Cefuroxime Axetil Tablets \\
J01BA01 & chloramphenicol & 5 & Chloramphenicol Injection \\
J01MA02 & ciprofloxacin & 8 & Ciprofloxacin and Sodium Chloride Injection \\
J01FA09 & clarithromycin & 4 & Clarithromycin Granule for Oral Suspension; Clarithromycin Tablets \\
J01FF01 & clindamycin & 4 & Clindamycin Phosphate for Injection \\
J01EE01 & compound sulfamethoxazole & 4 & Compound Sulfamethoxazole Tablets; Compound Sulfamethoxazole Injection \\
J01AA02 & doxycycline & 5 & --- \\
J01DH03 & erbepenem & 9 & Ertapenem for Injection \\
J01FA01 & erythromycin & 2 & Erythromycin Lactobionate for Injection; Erythromycin Ointment; Erythromycin Eye Ointment \\
 & extended spectrum beta-lactamase detection &  & --- \\
J02AC01 & fluconazole &  & Fluconazole Capsules; Fluconazole and Sodium Chloride Injection; Fluconazole Injection \\
J01MA16 & gatifloxacin & 9 & --- \\
J01GB03 & gentamicin & 5 & Gentamycin Sulfate and Procaine Hydrochloride Capsules; Gentamycin Sulfate,Procaine Hydrochloride and Vitamin B \\
J01GB03 & gentamicin 500 & 5 & Gentamycin Sulfate and Procaine Hydrochloride Capsules; Gentamycin Sulfate,Procaine Hydrochloride and Vitamin B \\
J01DH51 & imipenem & 10 & Imipenem and Cilastatin Sodium for Injection \\
 & induced clindamycin resistance &  & --- \\
J02AC02 & itraconazole &  & --- \\
J01FA07 & josamycin & 4 & --- \\
 & koalaranin &  & --- \\
J01MA12 & levofloxacin & 9 & Levofloxacin Eye Drops; Levofloxacin and Sodium Chloride Injection; Levofloxacin Hydrochloride Ear Drops \\
J01XX08 & linezolid & 5 & Linezolid Injection; Linezolid Tablets \\
J01DH02 & meropenem & 10 & Meropenem for Injection \\
 & methicillin-resistant staphylococcus &  & --- \\
J01AA08 & minocycline & 5 & --- \\
J01MA14 & moxifloxacin & 9 & --- \\
J01XE01 & nitrofurantoin & 2 & Nitrofurantoin Enteric-coated Tablets \\
J01MA01 & ofloxacin & 9 & Ofloxacin Eye Ointment; Ofloxacin Gel \\
J01CF04 & oxacillin & 1 & Oxacillin Sodium for Iniection \\
J01CE01 & penicillin & 2 & Benzylpenicillin Sodium for Injection; Benzathine Benzylpenicillin for Injection \\
J01CA12 & piperacillin & 5 & Piperacillin Sodium and Tazobactam Sodium Injection; Piperacillin Sodium and Tazobactam Sodium for Injection; Piperacillin Sodium And Sulbactam Sodium For Injection \\
J01CR05 & piperacillin/tazobactam & 8 & --- \\
J01XB02 & polymyxin b & 4 & Polymyxin B Sulfate for injection \\
J01FG02 & quinupristin/dalfopristin & 5 & --- \\
J04AB02 & rifampin & 3 & Rifampicin Capsules; Rifampicin for Eye Use \\
J01FA06 & roxithromycin & 4 & --- \\
J01MA09 & sparfloxacin & 8 & --- \\
J01GA01 & streptomycin 2000 & 5 & --- \\
J01FA15 & telithromycin & 4 & --- \\
J01AA07 & tetracycline & 5 & --- \\
J01CA13 & ticarcillin & 5 & --- \\
J01AA12 & tigecycline & 13 & Tigecycline for Injection \\
J01GB01 & tobramycin & 5 & Tobramycin and Dexamethasone Ophthalmic Ointment; Tobramycin Sulfate Injection; Tobramycin Dexamethasone Eye Drops; Tobramycin Eye Drops \\
J01XA01 & vancomycin & 5 & Vancomycin Hydrochloride for Intra Venous; Norvancomycin Hydrochloride for Injection \\
J02AC03 & voriconazole &  & Voriconazole for Injection; Voriconazole Tablets \\
 & $beta$-lactamase &  & --- \\

\end{longtable}


\section{Implementation details}
\label{app:methods_impl_details}

\subsection{Data preprocessing}
Numeric variables are clipped at the 1st and 99th percentile thresholds using Winsorization \citep{wilcoxIntroductionRobustEstimation2011} to reduce the impact of erroneous measurements and recording artifacts while preserving clinically plausible ranges. Undefined aggregates (e.g., no measurements within a window) are assigned neutral default values (e.g., zero counts), consistent with feature design. Data is further standardized using sklearn's \citep{scikit-learn} StandardScaler unless stated otherwise.

\subsection{Construction of antibiotic patient-day data}
The datasets are organized at the patient-day level: each row is one calendar day on the PICU with an overlapping antibiotic course. Antibiotic prescriptions within 24 hours are merged into clinically contiguous courses for each admission and antibiotic. Patient-day rows are enumerated only from those merged intervals. Multi-day antibiotic coverage therefore yields multiple rows; days without qualifying overlap are excluded. Merged 'same-day' antibiotic courses shorter than 24 hours are dropped as these are mostly prophylactic and the duration is too short for prospective audit and feedback. If several courses overlap the same day, all are considered when attaching flags (or across overlapping courses) rather than collapsing to a single agent. Prediction is defined at the start of each patient-day (DAY\_START), default 8:00AM, and all features are constructed using only information available up to this time point to prevent information leakage (DAY\_START-W,DAY\_START), with a default 24h aggregation window. 

\subsubsection{24h resolution}
Time-varying clinical variables, including vital signs, laboratory measurements, and treatment features, are aggregated within a fixed temporal windows anchored to the start of each patient-day. We compute clinically interpretable summary statistics (e.g., mean, extrema, and most recent value) to represent the current patient state. Treatment-related features capture antibiotic regimen characteristics such as route, number of concurrent agents, and days since antibiotic initiation. The final feature representation combines these components using one-hot encoding and engineered summary features, resulting in a total of 194 features.

\subsubsection{1 hour bins}
\label{app:impl_details_1h_bins}
We augment each patient-day timestep with a fixed-resolution grid of pre-anchor chart and laboratory measurements. For day index $d$ with anchor time \texttt{DAY\_START}(d), numeric vitals and labs with timestamps strictly in the leak-safe window [\texttt{DAY\_START}(d) - W,\ \texttt{DAY\_START}(d)) are aggregated to an hourly grid of 24 bins ($B$) per day. Empty bins are the forward-filled using LOCF \citep{engelsImputationMissingLongitudinal2003}, yielding a dense matrix of shape $(B \times S)$ with $S$ timeseries variables, as a simple multivariate representation of irregularly sampled vitals and labs on a common time base.

\paragraph{Encoding the temporal matrix.}
This grid is encoded by a small \texttt{EventGridEncoder} module comprising a linear projection per bin followed by two one-dimensional convolutions with global pooling over the bin axis, yielding a fixed-size embedding vector \texttt{event\_embed\_dim} per patient-day timestep. This separation avoids the sequence-length explosion that would result from unrolling hourly bins directly into the admission timeline (expanding sequence length by a factor of $B$).

\paragraph{Fusion of 1h embeddings to aggregated data for sequence models.}
The resulting \texttt{event\_embed\_dim} vector is concatenated with the scaled tabular feature vector for that day to form a fused representation of dimension \texttt{tab\_dim} + \texttt{event\_embed\_dim}, which is then passed through a LayerNorm before entering the sequence trunk and task heads. Where applicable, columns from the tabular feature vector whose information overlaps the longitudinal window were omitted from the tabular branch so coarse window-level summaries are not duplicated. The \texttt{EventGridEncoder} is not pre-trained or trained as a standalone classifier; it is optimized end-to-end within the full prediction network comprising the tabular pathway, event encoder, sequence trunk, and prediction head. The training objective is masked timestep binary cross-entropy with logits over valid sequence positions, with per-task positive-class reweighting, minimized with AdamW \citep{kingmaAdamMethodStochastic2017}. Gradients backpropagate through the prediction head, sequence trunk, fusion concatenation, and the Conv1d layers of the \texttt{EventGridEncoder}, so the convolutional filters learn whatever within-day temporal structure improves the final supervised prediction objective, subject to gradient clipping and early stopping on validation loss.

\paragraph{Fusion of 1h embeddings and aggregated data for tabular models.} 
For the tabular model [1h] benchmark, the same \texttt{EventGridEncoder} is trained with a lightweight per-day MLP trunk over the fused tabular and event embedding. The exported day-level event embeddings are then appended to the aggregated tabular feature representation and used by standard tabular classifiers. With this setup we can give the tabular models a fair temporal dimension, and understand if the difference between tabular and sequential models is not just because of access to more data.

\subsubsection{Sub-day irregular sequence representations} 
We collect the physiological timeseries data (Table~\ref{tab:extracted_data}) in the same leak-safe pre-anchor window used throughout the study, $[\texttt{DAY\_START}-W,\texttt{DAY\_START})$ with $W{=}24\,\mathrm{h}$. Rather than discretizing this window onto a fixed hourly grid and forward filling, we construct an irregular multivariate sequence whose timesteps are the unique observation times within the window. Each sample is represented by a value tensor, an observation mask, and a relative-time vector representing time from window start. This yields an irregular-time representation of EHR measurements that preserves asynchronous sampling and true missingness patterns within the day.

\subsection{Targets}
\label{app:targets}
For the outcomes we first derive course-level stewardship events from merged antibiotic prescription courses, then attach them to the patient-day unit. 

\paragraph{IV-to-oral.}
An IV-to-oral event is defined when a course contains an earlier intravenous prescription followed by a non-intravenous formulation. The timestamp of the first qualifying non-intravenous prescription is recorded as the event time. The corresponding patient-day is labeled positive only on the calendar day of the switch. Clinically, this target is best interpreted as a proxy for route-change review opportunity rather than a recommendation that a switch was appropriate.

\paragraph{De-escalation.}
A de-escalation event is defined when a subsequent antibiotic prescription within the same admission has a different Anatomic Therapeutic Classification (ATC) \citep{who_atc_2024} code and a strictly lower antibiotic spectrum index (ASI) \citep{gerber2017asi} than the current course (Table~\ref{tab:supp_antibiotic_spectrum}). Only the patient-day corresponding to the first qualifying de-escalation event is labeled positive. This target captures a prescription-based spectrum-narrowing proxy rather than adjudicated stewardship intent.

\paragraph{Discontinuation.}
A discontinuation event is defined when an antibiotic course ends before discharge, with at least 72 hours remaining in the admission and no overlapping antibiotic prescriptions in the subsequent interval. The label is assigned to the final day of the course. The label therefore depends on an operational drug-free-window threshold rather than direct clinician annotation.%

\paragraph{Short-course.}
A short-course event is defined when the duration of an antibiotic course is less than 96 hours. Note that since we exclude stays shorter than 24 hours, this means that a short-course antibiotic target is actually a course with a length between 24 and 96 hours. Additionally, Cefazolin (ATC: J01DB04) is excluded from this definition due to its common prophylactic use in PICU. As with discontinuation, only the final day of the course is labeled as positive. This is an auxiliary duration-based target and should likewise be viewed as a stewardship-relevant proxy rather than a gold-standard decision label.%

\subsection{Models}
\subsubsection{Tabular Models}
Tabular models operate on a flattened patient-day representation, where each instance corresponds to a single day of a patient’s stay. These models do not explicitly model temporal dependencies across timesteps. Instead, temporal information is incorporated indirectly through engineered features, such as length-of-stay variables and short-term trend summaries. As a result, each prediction is made independently based on the current patient state and aggregated context of the previous day, without access to the full temporal trajectory. We do not specifically handle class imbalance for the tabular models.

\paragraph{Logistic regression.}
Logistic regression uses L2 regularization with standard solver \texttt{lbfgs} and \texttt{max\_iter=3000} and default inverse of regularization strength \texttt{C=1.0})

\paragraph{LightGBM.}
LightGBM uses 500 trees with \texttt{learning\_rate=0.05} and \texttt{num\_leaves=31} \citep{ke2017lightgbm}. We don't standardize nor scale the features for this model. 

\paragraph{MLP classifier.}
Scikit-learn's \texttt{MLPClassifier} \citep{scikit-learn} was used with \texttt{hidden\_layer\_size=(64,0)}, ReLu activation function, batch\_size of 256, initial learning rate of 0.001, and 200 max iteration with early stopping.

\subsubsection{Tabular foundation model TabPFN}
TabPFN-v2.6 weights were downloaded locally from the official GitHub \url{https://github.com/PriorLabs/TabPFN} \citep{hollmann2025tabpfn}. Feature values were converted to floating point and non-finite values replaced with zeros before inference. No standardization or scaling is applied to the input data. For the \emph{MELODY-KISPI} cohort we trained/tested on the full sets (14,573/3,573 patient-days), for PIC we trained on the full training set (64,051), and tested on a stratified subset of 1000 patient because of compute limitations.

\subsubsection{Sequence-based temporal models}
Sequence-based temporal models train one network per target. Missing data are encoded using indicator variables in the sequence representations. Rows are grouped by admission id, ordered by timestamp of day start, truncated to at most 512 timesteps, and trained with masked per-timestep logistic losses so padded days do not contribute to optimization. Class imbalance is handled with a positive weight scalar for the target-specific loss, and the readout is one logit per patient-day rather than a pooled admission-level prediction.

\paragraph{Recurrent architectures.}
The recurrent models are Gated Recurrent Unit (GRU) \citep{cho2014learning} and Long Short-Term Memory (LSTM) \citep{hochreiter1997lstm} trunks with two stacked layers, hidden size 128, dropout 0.2 between recurrent layers, and a linear per-timestep head (unidirectional recurrent trunk).

\paragraph{Temporal Convolutional Network.}
The Temporal Convolutional Network (TCN) \citep{bai2018empirical} is implemented as a residual stack of left-padded causal dilated temporal convolutions over the admission trajectory. The causal structure ensures that predictions at timestep $t$ depend only on observations up to and including $t$. By using convolutions, the TCN is able to capture longer-range temporal dependencies while maintaining a fixed computational cost per timestep. The output at each timestep is passed to a linear prediction head.

\paragraph{Transformer.}
We use a Transformer encoder \citep{vaswani2017attention} with learned positional embeddings to model temporal dependencies. The architecture employs multi-head self-attention with GELU activations and a norm\_first configuration. To handle variable-length sequences, key-padding masks are applied to ignore padded timesteps. A causal attention mask is used to ensure that predictions at timestep $t$ depend only on observations up to and including $t$, preventing information leakage from future timesteps.

\paragraph{State Space Model.}
The state space model (SSM) is a lightweight Mamba-inspired selective SSM \citep{guMambaLinearTimeSequence2024}, implemented as a residual stack of gated, input-conditioned causal state-update blocks over the admission trajectory. Four input-dependent projections are computed from the hidden representation $\mathbf{x} \in \mathbb{R}^{B \times T \times H}$ at each timestep $t$. This gating mechanism allows the model to interpolate between propagating the updated latent state and passing the input through directly. Residual connections with dropout and a final layer normalisation are applied across the block stack, and padding safety is enforced by masking state updates and zeroing outputs beyond each sequence's valid length. The causal structure ensures that predictions at timestep $t$ depend only on observations up to and including $t$. The output at each timestep is passed to a linear prediction head.
This implementation is intentionally lightweight: it does not require a CUDA scan kernel and omits the convolutional mixer branch and full SSD parameterisation of the canonical Mamba release \citep{guMambaLinearTimeSequence2024}, prioritising ease of integration and fair comparison to the other simple sequence-based temporal architectures.

\subsubsection{Multi-task learning setting}
In the multi-task learning (MTL) setting of the sequence models, all four targets are predicted jointly using a shared encoder, with task-specific output heads. The models use the same per-timestep inputs as in the single task learning setting, i.e., tabular patient-day features alone, or fused with learnable features when applicable. We consider several MTL strategies. In hard parameter sharing, a single encoder is shared across tasks with separate linear heads. In uncertainty-weighted MTL \citep{kendall2018uncertainty}, task losses are combined using learned homoscedastic uncertainty parameters. We also evaluate a multi-gate mixture-of-experts (MMoE) architecture \citep{ma2018mmoe} with PCGrad \citep[projecting conflicting gradients]{yu2020pcgrad}, where the shared sequence representation is passed through a small set of expert networks with task-specific gating, and gradient updates are adjusted to reduce destructive interference between tasks. All MTL variants can be paired with GRU \citep{cho2014learning}, LSTM \citep{hochreiter1997lstm}, TCN \citep{bai2018empirical}, Transformer \citep{vaswani2017attention}, or SSM sequence encoders. 

\subsubsection{Sequence foundation model CLMBR-T-Base.}
\label{app:clmbr-t-base}

As a pretrained adult EHR sequence-model comparison, we used CLMBR-T-Base (Clinical Language-Model-Based Representations with a Transformer backbone), a 141M-parameter autoregressive model pretrained on structured EHR code sequences from approximately 2.57 million adult patients at Stanford Medicine \citep{wornow2023ehrshot}. CLMBR-T-Base was trained to predict the next medical code in a patient’s timeline given prior codes and produces fixed-dimensional patient representations intended for downstream prediction \citep{guo2024multi}. We used the publicly released CLMBR-T-Base checkpoint without fine-tuning the transformer weights.

\paragraph{Event representation and code mapping.}
For each hospital admission, we constructed a chronological sequence of coded clinical events in Medical Event Data Standard (MEDS, version 0.1.3) \citep{MedicalEventData2024}. This input schema is expected by the Framework for Electronic Medical Records (FEMR, version 0.2.3) \citep{femr2026} inference stack used to run CLMBR-T-Base. Events included date of birth, admission visit type, sex, time-stamped laboratory and vital-sign measurements, and antibiotic prescription starts. \\

Local PICU chart/lab item identifiers and antibiotic ATC codes were mapped to CLMBR-compatible vocabulary tokens (e.g., LOINC for measurements, RxNorm/ATC-derived drug codes for prescriptions, and visit/gender concept codes for static admission attributes). Unmapped local codes were omitted from each patient's timeline rather than imputed. We report the code mappings used below, which could contain errors as no specific specific documentation the e.g. admission types is provided by the \emph{PIC} database. Date of birth (DOB) was mapped to MEDS\_BIRTH, Sex to \textit{M} or \textit{F} for males and females, respectively. Because our PICU data only contains inpatients, we mapped the admission department to \textit{Visit/IP}. Further mappings for the admission type (Table~\ref{tab:clmbr-adm-type}), the measurement data (Table~\ref{tab:clmbr-itemid}), and the antibiotics (Table~\ref{tab:clmbr-atc5}) are specified below.

\begin{table}[H]
\centering
\caption{Mapping of Admission type to Visit.}
\label{tab:clmbr-adm-type}
\begin{tabular}{@{}ll@{}}
\toprule
\textbf{Admission type} & \textbf{Visit code} \\
\midrule
AMBULATORY OBSERVATION & Visit/OP \\
DIRECT OBSERVATION & Visit/IP \\
ELECTIVE & Visit/IP \\
EMERGENCY & Visit/ER \\
EU OBSERVATION & Visit/ER \\
NEWBORN & Visit/IP \\
OBSERVATION & Visit/OP \\
SURGICAL SAME DAY ADMISSION & Visit/OP \\
URGENT & Visit/IP \\
\_\_NA\_\_ & Visit/IP \\
\bottomrule
\end{tabular}
\end{table}

\begin{table}[H]
\centering
\captionsetup{width=0.6\textwidth}
\caption{Mapping of Chartevents and Labevents Item IDs to LOINC \citep{forreyLogicalObservationIdentifier1996}.}
\label{tab:clmbr-itemid}
\small
\begin{tabular}{@{}rlp{4cm}@{}}
\toprule
\textbf{Item ID} & \textbf{CLMBR code} & \textbf{Description} \\
\midrule
1001 & LOINC/8310-5 & Temperature \\
1003 & LOINC/8867-4 & Heart rate \\
1004 & LOINC/9279-1 & Respiratory rate \\
1006 & LOINC/59408-5 & Oxygen saturation \\
1013 & LOINC/8302-2 & Height \\
1014 & LOINC/29463-7 & Weight \\
1015 & LOINC/8462-4 & Diastolic BP \\
1016 & LOINC/8480-6 & Systolic BP \\
5024 & LOINC/1751-7 & Albumin \\
5026 & LOINC/1742-6 & ALT \\
5057 & LOINC/2532-0 & LDH \\
5099 & LOINC/718-7 & Hemoglobin \\
5132 & LOINC/789-8 & RBC \\
5141 & LOINC/6690-2 & WBC \\
5227 & LOINC/2524-7 & Lactate \\
5230 & LOINC/2947-0 & Sodium whole blood \\
5235 & LOINC/2019-8 & pCO2 ABG \\
5237 & LOINC/2746-6 & pH ABG \\
5239 & LOINC/2703-7 & pO2 ABG \\
5252 & LOINC/59408-5 & O2 sat blood gas \\
5626 & LOINC/1988-5 & CRP \\
6085 & LOINC/33959-8 & PCT \\
\bottomrule
\end{tabular}
\end{table}

{
\setlength{\LTcapwidth}{0.71\textwidth}
\begin{longtable}{@{}llp{5.5cm}@{}}
\caption{Antibiotics mapping of ATC \citep{who_atc_2024} to RxNorm \citep{rxnorm}.}
\label{tab:clmbr-atc5} \\
\toprule
\textbf{ATC} & \textbf{CLMBR code} & \textbf{Drug / notes} \\
\midrule
\endfirsthead
\multicolumn{3}{c}{\tablename\ \thetable\ -- \textit{Continued from previous page}} \\
\toprule
\textbf{ATC} & \textbf{CLMBR code} & \textbf{Drug / notes} \\
\midrule
\endhead
\midrule
\multicolumn{3}{r}{\textit{Continued on next page}} \\
\endfoot
\bottomrule
\endlastfoot
J01AA02 & RxNorm/3640 & doxycycline \\
J01CA04 & RxNorm/723 & Amoxicillin \\
J01CE01 & RxNorm/7980 & Benzathine penicillin \\
J01CE02 & RxNorm/7984 & penicillin V \\
J01CE08 & RxNorm/7984 & Penicillin G \\
J01CF01 & RxNorm/7987 & Dicloxacillin \\
J01CF04 & RxNorm/7989 & Oxacillin \\
J01CF05 & RxNorm/4448 & floxacillin \\
J01CG01 & RxNorm/10395 & Sulbactam \\
J01CR02 & RxNorm/1659141 & Amoxicillin-clavulanate \\
J01CR05 & RxNorm/1659139 & Piperacillin-tazobactam \\
J01DB01 & RxNorm/7984 & Penicillin (benzathine proxy) \\
J01DB04 & RxNorm/7984 & Penicillin G procaine \\
J01DC02 & RxNorm/1664986 & Cefazolin \\
J01DD01 & RxNorm/1665050 & Cefotaxime \\
J01DD02 & RxNorm/1665052 & Ceftazidime \\
J01DD04 & RxNorm/1665054 & Ceftriaxone \\
J01DD08 & RxNorm/1665056 & Cefepime \\
J01DE01 & RxNorm/1665060 & Meropenem \\
J01DH02 & RxNorm/1665062 & Imipenem \\
J01FA01 & RxNorm/4053 & erythromycin \\
J01FA09 & RxNorm/3640 & Clarithromycin \\
J01FA10 & RxNorm/3642 & Azithromycin \\
J01FF01 & RxNorm/3644 & Clindamycin \\
J01GB01 & RxNorm/10627 & tobramycin \\
J01GB03 & RxNorm/10395 & Gentamicin \\
J01GB06 & RxNorm/10396 & Amikacin \\
J01MA02 & RxNorm/7052 & Ciprofloxacin \\
J01MA12 & RxNorm/7054 & Levofloxacin \\
J01MA14 & RxNorm/7056 & Moxifloxacin \\
J01XA01 & RxNorm/11124 & Vancomycin \\
J01XA02 & RxNorm/57021 & teicoplanin \\
J01XD01 & RxNorm/7980 & Metronidazole \\
J01XE01 & RxNorm/7052 & Nitrofurantoin proxy \\
J01XX08 & RxNorm/11124 & Linezolid proxy \\
J02AA01 & RxNorm/732 & amphotericin B \\
J02AC01 & RxNorm/1908 & Fluconazole \\
J02AC02 & RxNorm/1909 & Itraconazole \\
J02AC03 & RxNorm/1910 & Voriconazole \\
J02AX04 & RxNorm/140108 & caspofungin \\
J04AB02 & RxNorm/9384 & rifampin \\
J04AC01 & RxNorm/6038 & isoniazid \\
J05AB01 & RxNorm/281 & acyclovir \\
J05AB06 & RxNorm/4678 & ganciclovir \\
J05AB11 & RxNorm/73645 & valacyclovir \\
J05AB12 & RxNorm/83171 & cidofovir \\
J05AB14 & RxNorm/275891 & valganciclovir \\
J05AB16 & RxNorm/2284718 & remdesivir \\
J05AD01 & RxNorm/33562 & foscarnet \\
J05AH02 & RxNorm/260101 & oseltamivir \\
\end{longtable}
}

\paragraph{Anchor-time representation extraction.}
Predictions were made at the same patient-day anchors used for all other models in our study: the start of each antibiotic stewardship day at \texttt{DAY\_START} (default 8:00AM). For each anchor, only events occurring strictly before that timestamp were included, ensuring a causal, leakage-free input window. For each admission, we ran one forward pass through the frozen CLMBR-T-Base model over the full preprocessed event timeline and extracted, at each anchor, the 768-dimensional hidden representation corresponding to the last model state at or before the anchor time. This yields one embedding vector per patient-day. We did not train a recurrent or temporal head across days; each day was scored from its anchor embedding alone.

\paragraph{Downstream prediction head.}
We fit a separate L2-regularized logistic regression model for each forward-looking stewardship target using the CLMBR-T-Base embeddings as the sole input features. Models used balanced class weights to mitigate label imbalance. 

\paragraph{Evaluation protocol.}
CLMBR-T-Base results were evaluated under the same patient-level temporal split and evaluated on the same fully held-out test set. CLMBR-T-Base was included as a reference baseline to assess how much signal an adult pretrained structured-EHR foundation model provides for PICU antibiotic stewardship tasks.

\subsubsection{Graph-based temporal model RAINDROP}
\label{app:RAINDROP}
We apply RAINDROP~\citep{Raindrop} to the sub-day irregular sequence representations to learn a fixed-dimensional temporal embedding for each patient-day. RAINDROP treats sensors as nodes of a graph and uses structured message passing to exchange information across variables at each observation time, thereby modeling inter-sensor dependencies. We use a standard RAINDROP configuration with 2 Transformer encoder layers, 2 attention heads, dropout of $0.2$, and 2 observation-propagation layers over the sensor graph. Temporal self-attention is applied causally over the irregular timeline, with time encoded via continuous positional features. The resulting per-timestep representations are aggregated by mean pooling over valid (non-padded) steps and concatenated with static tabular patient-context features. The fused vector is passed through the default two layer MLP classification head to produce binary predictions for each stewardship outcome. In this way, RAINDROP complements binned sequence models by operating directly on irregular observation times while remaining aligned with the same pre-anchor window.

\section{Additional Results}
\subsection{Cross-cohort prevalence}

\begin{table}[H]
\centering
\resizebox{0.7\textwidth}{!}{%
  \begin{tabular}{lcc}
\toprule
    \textbf{Quantity} & \textbf{PIC} & \textbf{MELODY-KISPI} \\
    \midrule
    Patient-days (all)   & 79,831 & 18,148  \\
    Patient-days (train) & 64,051 & 14,573  \\
    Patient-days (test)  & 15,780 & 3,575 \\
    Distinct patients (all)   & 5,515 & 2,059  \\
    Distinct admissions (all) & 5,671 & 2,789  \\
    \midrule
    \multicolumn{3}{l}{\textit{Test-set prevalence patient-days}} \\
    \quad IV-to-oral        & 0.0035 & 0.0098 \\
    \quad de-escalation     & 0.0392 & 0.0520 \\
    \quad discontinuation   & 0.0169 & 0.0568 \\
    \quad short-course      & 0.0577 & 0.1860  \\
    \midrule
    \multicolumn{3}{l}{\textit{Test-set positive-course duration, days (median [Q1, Q3])}} \\
    \quad IV-to-oral        & 6.62 [5.62, 9.25] & 3.19 [2.67, 5.07]  \\
    \quad de-escalation     & 6.62 [3.68, 10.02] & 1.99 [1.00, 3.30]  \\
    \quad discontinuation   & 6.76 [3.98, 9.71] & 2.00 [1.33, 4.33] \\
    \quad short-course      & 2.60 [1.81, 3.07] & 1.88 [1.01, 2.50]  \\
    \midrule
    \multicolumn{3}{l}{\textit{Age at treatment initiation, months (median [Q1, Q3])}} \\
    \quad Intravenous        & 10.8 [2.0, 48.4] & 5.0 [0.5, 42.3]  \\
    \quad Oral                & 14.6 [2.2, 60.4] & 8.7 [0.1, 26.8]  \\
    \bottomrule
\end{tabular}

}
\caption{Cross-cohort summary for \emph{PIC} and the \emph{MELODY-KISPI}. Test-set prevalences of the patient-day dataset and antibiotics course duration metrics split by targets are also included. We additionally report age metrics at treatment start.}\label{tab:cohort_summary}
\end{table}

\begin{figure}[H]
    \centering
    \includegraphics[width=1\linewidth]{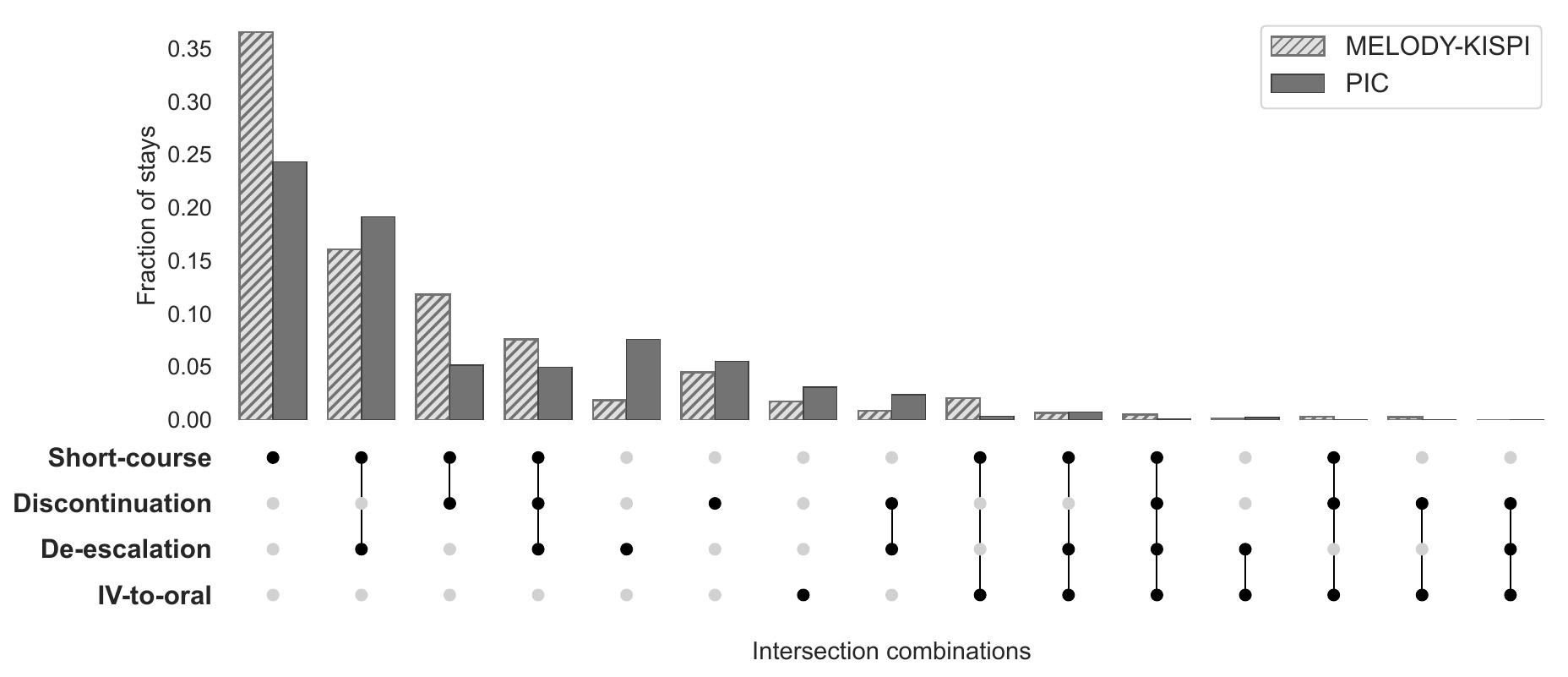}
    \caption{Intersections of AMS intervention targets on stay level across cohorts.}
    \label{fig:UpSet_overlap_outcomes}
\end{figure}

\subsection{Comparison to adults}
\label{app:comparison_to_adults}
\begin{table}[h]
    \centering
    \scriptsize
    \caption{Comparison of tabular models trained on 24-hour aggregated data with results reported by \cite{tran2024scaling} on a private Korean cohort and \cite{bolton2024personalising} on MIMIC-IV and eICU. For comparison, the early- and late-de-escalation target metrics of \cite{tran2024scaling} were averaged. We report AUROC, AUPRC, the true positive rate (TPR), true negative rate (TNR), positive predictive value (PPV), negative predictive value (NPV).}
    \label{tab:tran_the_bolton_comparison}
  \resizebox{\textwidth}{!}{\setlength{\tabcolsep}{3pt} 
{\tiny
\begin{tabular}{lllccccccc}
\toprule
Task & Cohort & Model & Prev. & AUROC & AUPRC & TPR & TNR & PPV & NPV \\
\midrule

IV-to-oral & MELODY-KISPI & LGBM & 0.010 & 0.808 & 0.033 & 0.000 & 1.000 & 0.000 & 0.992 \\
 &  & MLP & 0.010 & 0.555 & 0.009 & 0.083 & 0.845 & 0.004 & 0.991 \\
 & PIC & LGBM & 0.003 & 0.883 & 0.102 & 0.077 & 0.999 & 0.214 & 0.997 \\
 &  & MLP & 0.003 & 0.852 & 0.053 & 0.282 & 0.993 & 0.118 & 0.998 \\
& Tran The & LGBM & 0.030 & 0.810 & 0.160 & 0.780 & 0.710 & 0.070 & 0.990 \\
 \hdashline

& MIMIC-IV$^\dagger$ & MLP & -- & 0.80 & 0.37 & 0.85 & 0.25 & -- & -- \\
& eICU$^\dagger$ & MLP & -- & 0.77 & 0.33 & 0.90 & 0.35 & -- & -- \\
\midrule

De-escalation & MELODY-KISPI & LGBM & 0.052 & 0.932 & 0.396 & 0.147 & 0.991 & 0.490 & 0.952 \\
 &  & MLP & 0.052 & 0.873 & 0.272 & 0.552 & 0.906 & 0.256 & 0.972 \\
 & PIC & LGBM & 0.040 & 0.922 & 0.337 & 0.263 & 0.980 & 0.373 & 0.966 \\
 &  & MLP & 0.040 & 0.910 & 0.314 & 0.429 & 0.955 & 0.308 & 0.973 \\
& Tran The & LGBM & 0.035 & 0.750 & 0.130 & 0.640 & 0.760 & 0.080 & 0.985 \\
\midrule

Discontinuation & MELODY-KISPI & LGBM & 0.057 & 0.673 & 0.120 & 0.040 & 0.999 & 0.600 & 0.951 \\
 &  & MLP & 0.057 & 0.612 & 0.074 & 0.403 & 0.750 & 0.079 & 0.959 \\
 & PIC & LGBM & 0.017 & 0.835 & 0.110 & 0.063 & 0.998 & 0.293 & 0.986 \\
 &  & MLP & 0.017 & 0.816 & 0.066 & 0.137 & 0.982 & 0.102 & 0.987 \\
& Tran The & LGBM & 0.090 & 0.800 & 0.360 & 0.710 & 0.720 & 0.210 & 0.960 \\
\bottomrule
\end{tabular}
}
}
  \footnotesize{$^\dagger$ Not directly comparable to other cohorts due to differences in cohort selection and label definition (e.g., restricted to patients receiving both IV and oral antibiotics and evaluated at the per-day level).}
\end{table}

\clearpage
\subsection{Performance of models on finer resolution for MELODY-KISPI}
\label{app:perf_finer_resolution_private}
\begin{figure}[H]
    \centering
    \includegraphics[width=1\linewidth]{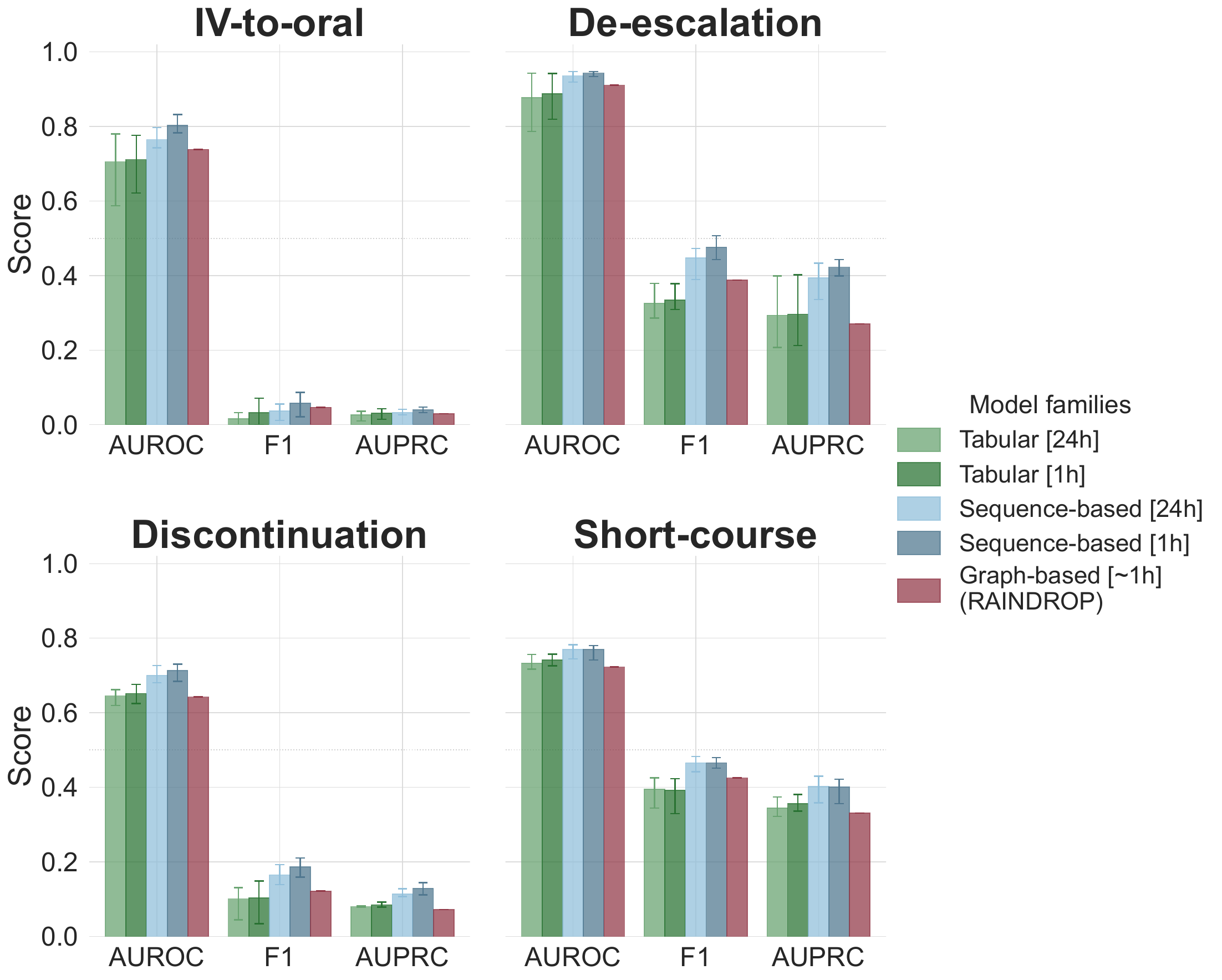}
    \caption{Performance of tabular and sequence-based models at 24-hour resolution and with augmented 1-hour representation, together with the graph-based model RAINDROP on irregular time series on the \emph{MELODY-KISPI} cohort. Bars and error bars show the model class mean and minimum-maximum range, respectively.}    
    \label{fig:Comparison_24h_1h_private}
\end{figure}

\clearpage
\subsection{Detailed results per target}
\label{app:detailed_results_targets}

\begin{table}[h!]
    \centering
    \scriptsize
    \caption{Full results for target IV-to-oral.}
    \label{tab:full_results_iv_to_oral}
  \resizebox{\textwidth}{!}{

\begin{tabular}{l*{10}{c}}
    \toprule
    & \multicolumn{10}{c}{{\large IV-to-oral}} \\
    \cmidrule(lr){2-11}
    & \multicolumn{5}{c}{MELODY-KISPI} & \multicolumn{5}{c}{PIC} \\
    \cmidrule(lr){2-6}\cmidrule(lr){7-11}
    \textbf{Model} & \textbf{AUROC} & \textbf{F1@0.5} & \textbf{F1@Threshold} & \textbf{Threshold} & \textbf{AUPRC} & \textbf{AUROC} & \textbf{F1@0.5} & \textbf{F1@Threshold} & \textbf{Threshold} & \textbf{AUPRC} \\
    \midrule
    \multicolumn{11}{l}{\textit{Tabular [24h]}} \\
    \cmidrule(lr){1-11}
    Logistic regression & 0.780 $\pm$ 0.000 & 0.000 $\pm$ 0.000 & 0.033 $\pm$ 0.000 & 0.078 & 0.034 $\pm$ 0.000 & 0.930 $\pm$ 0.000 & 0.000 $\pm$ 0.000 & 0.224 $\pm$ 0.000 & 0.089 & 0.137 $\pm$ 0.000 \\
    LightGBM & 0.749 $\pm$ 0.000 & 0.000 $\pm$ 0.000 & 0.000 $\pm$ 0.000 & 0.484 & 0.036 $\pm$ 0.000 & 0.942 $\pm$ 0.000 & 0.174 $\pm$ 0.000 & 0.159 $\pm$ 0.000 & 0.759 & 0.177 $\pm$ 0.000 \\
    MLP & 0.588 $\pm$ 0.051 & 0.000 $\pm$ 0.000 & 0.016 $\pm$ 0.010 & 0.088 & 0.010 $\pm$ 0.002 & 0.909 $\pm$ 0.012 & 0.000 $\pm$ 0.000 & 0.220 $\pm$ 0.035 & 0.071 & 0.120 $\pm$ 0.009 \\
    TabPFN & 0.810 $\pm$ 0.005 & 0.000 $\pm$ 0.000 & 0.030 $\pm$ 0.052 & 0.500 & 0.042 $\pm$ 0.002 & 0.978 $\pm$ 0.030 & 0.000 $\pm$ 0.000 & 0.000 $\pm$ 0.000 & 0.500 & 0.395 $\pm$ 0.096 \\
    \cmidrule(lr){1-11}
    \multicolumn{11}{l}{\textit{Tabular [1h]}} \\
    \cmidrule(lr){1-11}
    Logistic regression & 0.776 $\pm$ 0.000 & 0.000 $\pm$ 0.000 & 0.071 $\pm$ 0.000 & 0.066 & 0.034 $\pm$ 0.000 & 0.918 $\pm$ 0.000 & 0.000 $\pm$ 0.000 & 0.269 $\pm$ 0.000 & 0.097 & 0.144 $\pm$ 0.000 \\
    LightGBM & 0.737 $\pm$ 0.000 & 0.000 $\pm$ 0.000 & 0.000 $\pm$ 0.000 & 0.489 & 0.043 $\pm$ 0.000 & 0.930 $\pm$ 0.000 & 0.222 $\pm$ 0.000 & 0.100 $\pm$ 0.000 & 0.876 & 0.172 $\pm$ 0.000 \\
    MLP & 0.621 $\pm$ 0.040 & 0.000 $\pm$ 0.000 & 0.027 $\pm$ 0.001 & 0.039 & 0.015 $\pm$ 0.001 & 0.879 $\pm$ 0.021 & 0.000 $\pm$ 0.000 & 0.184 $\pm$ 0.026 & 0.069 & 0.091 $\pm$ 0.013 \\
    \cmidrule(lr){1-11}
    \multicolumn{11}{l}{\textit{Sequence-based [24h]}} \\
    \cmidrule(lr){1-11}
    GRU & 0.745 $\pm$ 0.001 & 0.049 $\pm$ 0.003 & 0.028 $\pm$ 0.026 & 0.947 & 0.027 $\pm$ 0.003 & 0.949 $\pm$ 0.007 & 0.074 $\pm$ 0.007 & 0.182 $\pm$ 0.009 & 0.965 & 0.093 $\pm$ 0.002 \\
    LSTM & 0.747 $\pm$ 0.011 & 0.046 $\pm$ 0.003 & 0.047 $\pm$ 0.006 & 0.897 & 0.028 $\pm$ 0.002 & 0.931 $\pm$ 0.011 & 0.075 $\pm$ 0.005 & 0.217 $\pm$ 0.035 & 0.962 & 0.116 $\pm$ 0.011 \\
    TCN & 0.743 $\pm$ 0.002 & 0.050 $\pm$ 0.008 & 0.012 $\pm$ 0.020 & 0.930 & 0.027 $\pm$ 0.001 & 0.954 $\pm$ 0.002 & 0.081 $\pm$ 0.011 & 0.202 $\pm$ 0.028 & 0.955 & 0.129 $\pm$ 0.032 \\
    Transformer & 0.794 $\pm$ 0.025 & 0.051 $\pm$ 0.005 & 0.056 $\pm$ 0.049 & 0.952 & 0.041 $\pm$ 0.007 & 0.930 $\pm$ 0.012 & 0.093 $\pm$ 0.024 & 0.262 $\pm$ 0.034 & 0.992 & 0.167 $\pm$ 0.047 \\
    SSM & 0.797 $\pm$ 0.012 & 0.060 $\pm$ 0.005 & 0.041 $\pm$ 0.037 & 0.876 & 0.038 $\pm$ 0.006 & 0.951 $\pm$ 0.006 & 0.089 $\pm$ 0.002 & 0.182 $\pm$ 0.031 & 0.965 & 0.112 $\pm$ 0.023 \\
    CLMBR-T-base & 0.723 $\pm$ 0.029 & 0.053 $\pm$ 0.002 & 0.036 $\pm$ 0.031 & 0.692 & 0.022 $\pm$ 0.002 & 0.877 $\pm$ 0.003 & 0.087 $\pm$ 0.010 & 0.114 $\pm$ 0.011 & 0.983 & 0.049 $\pm$ 0.002 \\
    \cmidrule(lr){1-11}
    \multicolumn{11}{l}{\textit{Sequence-based [1h]}} \\
    \cmidrule(lr){1-11}
    GRU & 0.789 $\pm$ 0.007 & 0.050 $\pm$ 0.008 & 0.062 $\pm$ 0.031 & 0.908 & 0.032 $\pm$ 0.005 & 0.946 $\pm$ 0.004 & 0.082 $\pm$ 0.011 & 0.205 $\pm$ 0.024 & 0.981 & 0.106 $\pm$ 0.017 \\
    LSTM & 0.783 $\pm$ 0.003 & 0.056 $\pm$ 0.002 & 0.042 $\pm$ 0.016 & 0.908 & 0.037 $\pm$ 0.007 & 0.924 $\pm$ 0.036 & 0.091 $\pm$ 0.028 & 0.193 $\pm$ 0.011 & 0.973 & 0.151 $\pm$ 0.004 \\
    TCN & 0.800 $\pm$ 0.013 & 0.050 $\pm$ 0.004 & 0.022 $\pm$ 0.037 & 0.936 & 0.036 $\pm$ 0.007 & 0.949 $\pm$ 0.007 & 0.098 $\pm$ 0.003 & 0.204 $\pm$ 0.003 & 0.967 & 0.135 $\pm$ 0.033 \\
    Transformer & 0.814 $\pm$ 0.002 & 0.057 $\pm$ 0.008 & 0.076 $\pm$ 0.018 & 0.936 & 0.044 $\pm$ 0.007 & 0.929 $\pm$ 0.005 & 0.093 $\pm$ 0.022 & 0.249 $\pm$ 0.028 & 0.993 & 0.155 $\pm$ 0.005 \\
    SSM & 0.832 $\pm$ 0.009 & 0.052 $\pm$ 0.006 & 0.087 $\pm$ 0.009 & 0.862 & 0.048 $\pm$ 0.003 & 0.942 $\pm$ 0.011 & 0.084 $\pm$ 0.003 & 0.216 $\pm$ 0.025 & 0.993 & 0.133 $\pm$ 0.013 \\
    \cmidrule(lr){1-11}
    \multicolumn{11}{l}{\textit{Graph [\textasciitilde{}1h]}} \\
    \cmidrule(lr){1-11}
    RAINDROP & 0.738 $\pm$ 0.015 & 0.011 $\pm$ 0.019 & 0.047 $\pm$ 0.029 & 0.288 & 0.030 $\pm$ 0.002 & 0.908 $\pm$ 0.015 & 0.070 $\pm$ 0.121 & 0.171 $\pm$ 0.023 & 0.296 & 0.119 $\pm$ 0.012 \\
    \bottomrule
\end{tabular}
}
\end{table}

\begin{table}[h!]
    \centering
    \scriptsize
    \caption{Full results for target De-escalation.}
    \label{tab:full_results_deescalation}
  \resizebox{\textwidth}{!}{

\begin{tabular}{l*{10}{c}}
    \toprule
    & \multicolumn{10}{c}{{\large De-escalation}} \\
    \cmidrule(lr){2-11}
    & \multicolumn{5}{c}{MELODY-KISPI} & \multicolumn{5}{c}{PIC} \\
    \cmidrule(lr){2-6}\cmidrule(lr){7-11}
    \textbf{Model} & \textbf{AUROC} & \textbf{F1@0.5} & \textbf{F1@Threshold} & \textbf{Threshold} & \textbf{AUPRC} & \textbf{AUROC} & \textbf{F1@0.5} & \textbf{F1@Threshold} & \textbf{Threshold} & \textbf{AUPRC} \\
    \midrule
    \multicolumn{11}{l}{\textit{Tabular [24h]}} \\
    \cmidrule(lr){1-11}
    Logistic regression & 0.903 $\pm$ 0.000 & 0.159 $\pm$ 0.000 & 0.379 $\pm$ 0.000 & 0.108 & 0.275 $\pm$ 0.000 & 0.910 $\pm$ 0.000 & 0.177 $\pm$ 0.000 & 0.356 $\pm$ 0.000 & 0.123 & 0.264 $\pm$ 0.000 \\
    LightGBM & 0.943 $\pm$ 0.000 & 0.352 $\pm$ 0.000 & 0.312 $\pm$ 0.000 & 0.555 & 0.399 $\pm$ 0.000 & 0.936 $\pm$ 0.000 & 0.293 $\pm$ 0.000 & 0.390 $\pm$ 0.000 & 0.315 & 0.410 $\pm$ 0.000 \\
    MLP & 0.786 $\pm$ 0.059 & 0.055 $\pm$ 0.050 & 0.286 $\pm$ 0.097 & 0.178 & 0.207 $\pm$ 0.065 & 0.917 $\pm$ 0.005 & 0.237 $\pm$ 0.044 & 0.354 $\pm$ 0.015 & 0.209 & 0.304 $\pm$ 0.006 \\
    TabPFN & 0.952 $\pm$ 0.001 & 0.378 $\pm$ 0.027 & 0.430 $\pm$ 0.072 & 0.500 & 0.479 $\pm$ 0.006 & 0.942 $\pm$ 0.024 & 0.222 $\pm$ 0.071 & 0.222 $\pm$ 0.071 & 0.500 & 0.469 $\pm$ 0.137 \\
    \cmidrule(lr){1-11}
    \multicolumn{11}{l}{\textit{Tabular [1h]}} \\
    \cmidrule(lr){1-11}
    Logistic regression & 0.904 $\pm$ 0.000 & 0.160 $\pm$ 0.000 & 0.378 $\pm$ 0.000 & 0.108 & 0.275 $\pm$ 0.000 & 0.909 $\pm$ 0.000 & 0.155 $\pm$ 0.000 & 0.359 $\pm$ 0.000 & 0.119 & 0.260 $\pm$ 0.000 \\
    LightGBM & 0.942 $\pm$ 0.000 & 0.356 $\pm$ 0.000 & 0.309 $\pm$ 0.000 & 0.545 & 0.402 $\pm$ 0.000 & 0.934 $\pm$ 0.000 & 0.283 $\pm$ 0.000 & 0.378 $\pm$ 0.000 & 0.300 & 0.387 $\pm$ 0.000 \\
    MLP & 0.820 $\pm$ 0.085 & 0.013 $\pm$ 0.023 & 0.315 $\pm$ 0.052 & 0.110 & 0.213 $\pm$ 0.059 & 0.915 $\pm$ 0.014 & 0.180 $\pm$ 0.065 & 0.352 $\pm$ 0.022 & 0.196 & 0.279 $\pm$ 0.034 \\
    \cmidrule(lr){1-11}
    \multicolumn{11}{l}{\textit{Sequence-based [24h]}} \\
    \cmidrule(lr){1-11}
    GRU & 0.948 $\pm$ 0.001 & 0.428 $\pm$ 0.013 & 0.467 $\pm$ 0.017 & 0.817 & 0.433 $\pm$ 0.033 & 0.962 $\pm$ 0.001 & 0.423 $\pm$ 0.026 & 0.524 $\pm$ 0.014 & 0.908 & 0.524 $\pm$ 0.015 \\
    LSTM & 0.941 $\pm$ 0.005 & 0.399 $\pm$ 0.035 & 0.473 $\pm$ 0.038 & 0.823 & 0.415 $\pm$ 0.040 & 0.962 $\pm$ 0.002 & 0.409 $\pm$ 0.022 & 0.532 $\pm$ 0.014 & 0.877 & 0.519 $\pm$ 0.016 \\
    TCN & 0.941 $\pm$ 0.008 & 0.391 $\pm$ 0.032 & 0.466 $\pm$ 0.015 & 0.842 & 0.392 $\pm$ 0.017 & 0.961 $\pm$ 0.004 & 0.382 $\pm$ 0.029 & 0.524 $\pm$ 0.030 & 0.866 & 0.483 $\pm$ 0.025 \\
    Transformer & 0.919 $\pm$ 0.027 & 0.310 $\pm$ 0.062 & 0.390 $\pm$ 0.083 & 0.807 & 0.336 $\pm$ 0.088 & 0.956 $\pm$ 0.000 & 0.352 $\pm$ 0.016 & 0.473 $\pm$ 0.011 & 0.885 & 0.437 $\pm$ 0.013 \\
    SSM & 0.932 $\pm$ 0.008 & 0.360 $\pm$ 0.007 & 0.442 $\pm$ 0.031 & 0.858 & 0.395 $\pm$ 0.035 & 0.959 $\pm$ 0.001 & 0.349 $\pm$ 0.006 & 0.496 $\pm$ 0.011 & 0.914 & 0.477 $\pm$ 0.008 \\
    CLMBR-T-base & 0.585 $\pm$ 0.004 & 0.110 $\pm$ 0.001 & 0.120 $\pm$ 0.016 & 0.576 & 0.088 $\pm$ 0.007 & 0.686 $\pm$ 0.001 & 0.122 $\pm$ 0.001 & 0.129 $\pm$ 0.005 & 0.738 & 0.070 $\pm$ 0.001 \\
    \cmidrule(lr){1-11}
    \multicolumn{11}{l}{\textit{Sequence-based [1h]}} \\
    \cmidrule(lr){1-11}
    GRU & 0.947 $\pm$ 0.003 & 0.411 $\pm$ 0.008 & 0.508 $\pm$ 0.035 & 0.885 & 0.441 $\pm$ 0.051 & 0.966 $\pm$ 0.002 & 0.397 $\pm$ 0.024 & 0.536 $\pm$ 0.011 & 0.891 & 0.543 $\pm$ 0.015 \\
    LSTM & 0.946 $\pm$ 0.002 & 0.393 $\pm$ 0.016 & 0.487 $\pm$ 0.006 & 0.902 & 0.443 $\pm$ 0.037 & 0.965 $\pm$ 0.002 & 0.391 $\pm$ 0.020 & 0.533 $\pm$ 0.012 & 0.869 & 0.537 $\pm$ 0.019 \\
    TCN & 0.946 $\pm$ 0.004 & 0.406 $\pm$ 0.024 & 0.480 $\pm$ 0.022 & 0.862 & 0.418 $\pm$ 0.038 & 0.965 $\pm$ 0.002 & 0.381 $\pm$ 0.004 & 0.530 $\pm$ 0.017 & 0.871 & 0.524 $\pm$ 0.021 \\
    Transformer & 0.934 $\pm$ 0.004 & 0.295 $\pm$ 0.027 & 0.443 $\pm$ 0.043 & 0.850 & 0.413 $\pm$ 0.050 & 0.955 $\pm$ 0.003 & 0.335 $\pm$ 0.035 & 0.469 $\pm$ 0.010 & 0.870 & 0.443 $\pm$ 0.039 \\
    SSM & 0.944 $\pm$ 0.001 & 0.376 $\pm$ 0.001 & 0.463 $\pm$ 0.017 & 0.883 & 0.399 $\pm$ 0.017 & 0.962 $\pm$ 0.002 & 0.367 $\pm$ 0.032 & 0.508 $\pm$ 0.006 & 0.933 & 0.504 $\pm$ 0.007 \\
    \cmidrule(lr){1-11}
    \multicolumn{11}{l}{\textit{Graph [\textasciitilde{}1h]}} \\
    \cmidrule(lr){1-11}
    RAINDROP & 0.907 $\pm$ 0.025 & 0.383 $\pm$ 0.025 & 0.388 $\pm$ 0.027 & 0.557 & 0.273 $\pm$ 0.008 & 0.896 $\pm$ 0.005 & 0.308 $\pm$ 0.010 & 0.331 $\pm$ 0.007 & 0.669 & 0.233 $\pm$ 0.007 \\
    \bottomrule
\end{tabular}
}
\end{table}

\begin{table}[h!]
    \centering
    \scriptsize
    \caption{Full results for target Discontinuation.}
    \label{tab:full_results_discontinuation}
  \resizebox{\textwidth}{!}{

\begin{tabular}{l*{10}{c}}
    \toprule
    & \multicolumn{10}{c}{{\large Discontinuation}} \\
    \cmidrule(lr){2-11}
    & \multicolumn{5}{c}{MELODY-KISPI} & \multicolumn{5}{c}{PIC} \\
    \cmidrule(lr){2-6}\cmidrule(lr){7-11}
    \textbf{Model} & \textbf{AUROC} & \textbf{F1@0.5} & \textbf{F1@Threshold} & \textbf{Threshold} & \textbf{AUPRC} & \textbf{AUROC} & \textbf{F1@0.5} & \textbf{F1@Threshold} & \textbf{Threshold} & \textbf{AUPRC} \\
    \midrule
    \multicolumn{11}{l}{\textit{Tabular [24h]}} \\
    \cmidrule(lr){1-11}
    Logistic regression & 0.654 $\pm$ 0.000 & 0.000 $\pm$ 0.000 & 0.131 $\pm$ 0.000 & 0.123 & 0.080 $\pm$ 0.000 & 0.796 $\pm$ 0.000 & 0.000 $\pm$ 0.000 & 0.119 $\pm$ 0.000 & 0.054 & 0.060 $\pm$ 0.000 \\
    LightGBM & 0.662 $\pm$ 0.000 & 0.000 $\pm$ 0.000 & 0.044 $\pm$ 0.000 & 0.289 & 0.082 $\pm$ 0.000 & 0.841 $\pm$ 0.000 & 0.014 $\pm$ 0.000 & 0.098 $\pm$ 0.000 & 0.188 & 0.097 $\pm$ 0.000 \\
    MLP & 0.620 $\pm$ 0.023 & 0.000 $\pm$ 0.000 & 0.128 $\pm$ 0.015 & 0.102 & 0.079 $\pm$ 0.007 & 0.740 $\pm$ 0.020 & 0.000 $\pm$ 0.000 & 0.105 $\pm$ 0.007 & 0.065 & 0.053 $\pm$ 0.003 \\
    TabPFN & 0.721 $\pm$ 0.004 & 0.000 $\pm$ 0.000 & 0.116 $\pm$ 0.102 & 0.500 & 0.122 $\pm$ 0.006 & 0.903 $\pm$ 0.025 & 0.000 $\pm$ 0.000 & 0.000 $\pm$ 0.000 & 0.500 & 0.278 $\pm$ 0.093 \\
    \cmidrule(lr){1-11}
    \multicolumn{11}{l}{\textit{Tabular [1h]}} \\
    \cmidrule(lr){1-11}
    Logistic regression & 0.654 $\pm$ 0.000 & 0.000 $\pm$ 0.000 & 0.149 $\pm$ 0.000 & 0.109 & 0.083 $\pm$ 0.000 & 0.775 $\pm$ 0.000 & 0.000 $\pm$ 0.000 & 0.112 $\pm$ 0.000 & 0.051 & 0.052 $\pm$ 0.000 \\
    LightGBM & 0.676 $\pm$ 0.000 & 0.000 $\pm$ 0.000 & 0.034 $\pm$ 0.000 & 0.281 & 0.092 $\pm$ 0.000 & 0.828 $\pm$ 0.000 & 0.014 $\pm$ 0.000 & 0.117 $\pm$ 0.000 & 0.168 & 0.112 $\pm$ 0.000 \\
    MLP & 0.625 $\pm$ 0.013 & 0.000 $\pm$ 0.000 & 0.128 $\pm$ 0.011 & 0.067 & 0.078 $\pm$ 0.007 & 0.744 $\pm$ 0.034 & 0.000 $\pm$ 0.000 & 0.110 $\pm$ 0.012 & 0.058 & 0.057 $\pm$ 0.008 \\
    \cmidrule(lr){1-11}
    \multicolumn{11}{l}{\textit{Sequence-based [24h]}} \\
    \cmidrule(lr){1-11}
    GRU & 0.726 $\pm$ 0.004 & 0.159 $\pm$ 0.003 & 0.193 $\pm$ 0.015 & 0.743 & 0.128 $\pm$ 0.007 & 0.835 $\pm$ 0.009 & 0.095 $\pm$ 0.004 & 0.151 $\pm$ 0.012 & 0.857 & 0.093 $\pm$ 0.011 \\
    LSTM & 0.704 $\pm$ 0.003 & 0.154 $\pm$ 0.010 & 0.156 $\pm$ 0.009 & 0.731 & 0.107 $\pm$ 0.010 & 0.836 $\pm$ 0.008 & 0.094 $\pm$ 0.004 & 0.164 $\pm$ 0.001 & 0.889 & 0.105 $\pm$ 0.004 \\
    TCN & 0.706 $\pm$ 0.011 & 0.159 $\pm$ 0.004 & 0.157 $\pm$ 0.016 & 0.706 & 0.113 $\pm$ 0.015 & 0.819 $\pm$ 0.015 & 0.083 $\pm$ 0.003 & 0.127 $\pm$ 0.019 & 0.814 & 0.068 $\pm$ 0.011 \\
    Transformer & 0.685 $\pm$ 0.011 & 0.140 $\pm$ 0.004 & 0.139 $\pm$ 0.007 & 0.663 & 0.107 $\pm$ 0.009 & 0.823 $\pm$ 0.011 & 0.087 $\pm$ 0.003 & 0.135 $\pm$ 0.012 & 0.857 & 0.081 $\pm$ 0.008 \\
    SSM & 0.680 $\pm$ 0.018 & 0.140 $\pm$ 0.010 & 0.174 $\pm$ 0.022 & 0.701 & 0.111 $\pm$ 0.005 & 0.809 $\pm$ 0.002 & 0.086 $\pm$ 0.002 & 0.124 $\pm$ 0.001 & 0.826 & 0.067 $\pm$ 0.003 \\
    CLMBR-T-base & 0.614 $\pm$ 0.006 & 0.135 $\pm$ 0.009 & 0.122 $\pm$ 0.007 & 0.544 & 0.077 $\pm$ 0.002 & 0.720 $\pm$ 0.002 & 0.071 $\pm$ 0.001 & 0.083 $\pm$ 0.009 & 0.863 & 0.041 $\pm$ 0.000 \\
    \cmidrule(lr){1-11}
    \multicolumn{11}{l}{\textit{Sequence-based [1h]}} \\
    \cmidrule(lr){1-11}
    GRU & 0.728 $\pm$ 0.010 & 0.160 $\pm$ 0.005 & 0.191 $\pm$ 0.033 & 0.742 & 0.132 $\pm$ 0.020 & 0.824 $\pm$ 0.011 & 0.086 $\pm$ 0.009 & 0.141 $\pm$ 0.014 & 0.850 & 0.081 $\pm$ 0.007 \\
    LSTM & 0.723 $\pm$ 0.016 & 0.155 $\pm$ 0.005 & 0.178 $\pm$ 0.009 & 0.690 & 0.137 $\pm$ 0.012 & 0.823 $\pm$ 0.005 & 0.085 $\pm$ 0.002 & 0.147 $\pm$ 0.007 & 0.831 & 0.085 $\pm$ 0.003 \\
    TCN & 0.731 $\pm$ 0.018 & 0.163 $\pm$ 0.008 & 0.211 $\pm$ 0.010 & 0.720 & 0.144 $\pm$ 0.015 & 0.826 $\pm$ 0.009 & 0.087 $\pm$ 0.008 & 0.149 $\pm$ 0.009 & 0.855 & 0.081 $\pm$ 0.012 \\
    Transformer & 0.684 $\pm$ 0.011 & 0.145 $\pm$ 0.003 & 0.159 $\pm$ 0.012 & 0.592 & 0.111 $\pm$ 0.007 & 0.808 $\pm$ 0.007 & 0.072 $\pm$ 0.006 & 0.132 $\pm$ 0.012 & 0.878 & 0.073 $\pm$ 0.006 \\
    SSM & 0.699 $\pm$ 0.012 & 0.145 $\pm$ 0.012 & 0.196 $\pm$ 0.020 & 0.706 & 0.119 $\pm$ 0.009 & 0.803 $\pm$ 0.006 & 0.086 $\pm$ 0.005 & 0.133 $\pm$ 0.017 & 0.856 & 0.076 $\pm$ 0.007 \\
    \cmidrule(lr){1-11}
    \multicolumn{11}{l}{\textit{Graph [\textasciitilde{}1h]}} \\
    \cmidrule(lr){1-11}
    RAINDROP & 0.627 $\pm$ 0.026 & 0.130 $\pm$ 0.007 & 0.109 $\pm$ 0.033 & 0.520 & 0.069 $\pm$ 0.007 & 0.739 $\pm$ 0.007 & 0.074 $\pm$ 0.019 & 0.094 $\pm$ 0.004 & 0.379 & 0.047 $\pm$ 0.001 \\
    \bottomrule
\end{tabular}
}
\end{table}

\begin{table}[h!]
    \centering
    \scriptsize
    \caption{Full results for target Short-course.}
    \label{tab:full_results_short_course}
  \resizebox{\textwidth}{!}{

\begin{tabular}{l*{10}{c}}
    \toprule
    & \multicolumn{10}{c}{{\large Short-course}} \\
    \cmidrule(lr){2-11}
    & \multicolumn{5}{c}{MELODY-KISPI} & \multicolumn{5}{c}{PIC} \\
    \cmidrule(lr){2-6}\cmidrule(lr){7-11}
    \textbf{Model} & \textbf{AUROC} & \textbf{F1@0.5} & \textbf{F1@Threshold} & \textbf{Threshold} & \textbf{AUPRC} & \textbf{AUROC} & \textbf{F1@0.5} & \textbf{F1@Threshold} & \textbf{Threshold} & \textbf{AUPRC} \\
    \midrule
    \multicolumn{11}{l}{\textit{Tabular [24h]}} \\
    \cmidrule(lr){1-11}
    Logistic regression & 0.726 $\pm$ 0.000 & 0.048 $\pm$ 0.000 & 0.425 $\pm$ 0.000 & 0.185 & 0.337 $\pm$ 0.000 & 0.745 $\pm$ 0.000 & 0.034 $\pm$ 0.000 & 0.216 $\pm$ 0.000 & 0.141 & 0.152 $\pm$ 0.000 \\
    LightGBM & 0.756 $\pm$ 0.000 & 0.193 $\pm$ 0.000 & 0.345 $\pm$ 0.000 & 0.358 & 0.374 $\pm$ 0.000 & 0.805 $\pm$ 0.000 & 0.183 $\pm$ 0.000 & 0.326 $\pm$ 0.000 & 0.224 & 0.287 $\pm$ 0.000 \\
    MLP & 0.717 $\pm$ 0.026 & 0.086 $\pm$ 0.059 & 0.414 $\pm$ 0.028 & 0.220 & 0.322 $\pm$ 0.028 & 0.766 $\pm$ 0.006 & 0.133 $\pm$ 0.018 & 0.257 $\pm$ 0.009 & 0.181 & 0.194 $\pm$ 0.009 \\
    TabPFN & 0.779 $\pm$ 0.001 & 0.144 $\pm$ 0.010 & 0.363 $\pm$ 0.188 & 0.500 & 0.412 $\pm$ 0.003 & 0.840 $\pm$ 0.007 & 0.173 $\pm$ 0.108 & 0.173 $\pm$ 0.108 & 0.500 & 0.342 $\pm$ 0.062 \\
    \cmidrule(lr){1-11}
    \multicolumn{11}{l}{\textit{Tabular [1h]}} \\
    \cmidrule(lr){1-11}
    Logistic regression & 0.726 $\pm$ 0.000 & 0.060 $\pm$ 0.000 & 0.422 $\pm$ 0.000 & 0.196 & 0.336 $\pm$ 0.000 & 0.742 $\pm$ 0.000 & 0.030 $\pm$ 0.000 & 0.218 $\pm$ 0.000 & 0.149 & 0.149 $\pm$ 0.000 \\
    LightGBM & 0.757 $\pm$ 0.000 & 0.193 $\pm$ 0.000 & 0.329 $\pm$ 0.000 & 0.371 & 0.381 $\pm$ 0.000 & 0.794 $\pm$ 0.000 & 0.181 $\pm$ 0.000 & 0.300 $\pm$ 0.000 & 0.247 & 0.272 $\pm$ 0.000 \\
    MLP & 0.742 $\pm$ 0.005 & 0.121 $\pm$ 0.022 & 0.423 $\pm$ 0.010 & 0.236 & 0.352 $\pm$ 0.002 & 0.766 $\pm$ 0.003 & 0.138 $\pm$ 0.019 & 0.263 $\pm$ 0.004 & 0.179 & 0.195 $\pm$ 0.014 \\
    \cmidrule(lr){1-11}
    \multicolumn{11}{l}{\textit{Sequence-based [24h]}} \\
    \cmidrule(lr){1-11}
    GRU & 0.776 $\pm$ 0.003 & 0.471 $\pm$ 0.002 & 0.471 $\pm$ 0.018 & 0.584 & 0.411 $\pm$ 0.006 & 0.848 $\pm$ 0.003 & 0.249 $\pm$ 0.008 & 0.365 $\pm$ 0.005 & 0.859 & 0.343 $\pm$ 0.013 \\
    LSTM & 0.780 $\pm$ 0.006 & 0.471 $\pm$ 0.004 & 0.483 $\pm$ 0.006 & 0.622 & 0.419 $\pm$ 0.010 & 0.832 $\pm$ 0.005 & 0.236 $\pm$ 0.007 & 0.356 $\pm$ 0.005 & 0.823 & 0.332 $\pm$ 0.011 \\
    TCN & 0.783 $\pm$ 0.011 & 0.473 $\pm$ 0.001 & 0.465 $\pm$ 0.018 & 0.588 & 0.430 $\pm$ 0.021 & 0.828 $\pm$ 0.006 & 0.247 $\pm$ 0.019 & 0.343 $\pm$ 0.005 & 0.797 & 0.286 $\pm$ 0.005 \\
    Transformer & 0.768 $\pm$ 0.001 & 0.468 $\pm$ 0.008 & 0.459 $\pm$ 0.015 & 0.578 & 0.394 $\pm$ 0.008 & 0.808 $\pm$ 0.005 & 0.228 $\pm$ 0.005 & 0.289 $\pm$ 0.012 & 0.809 & 0.252 $\pm$ 0.013 \\
    SSM & 0.744 $\pm$ 0.003 & 0.444 $\pm$ 0.006 & 0.442 $\pm$ 0.007 & 0.561 & 0.359 $\pm$ 0.002 & 0.769 $\pm$ 0.005 & 0.207 $\pm$ 0.006 & 0.260 $\pm$ 0.017 & 0.788 & 0.204 $\pm$ 0.020 \\
    CLMBR-T-base & 0.650 $\pm$ 0.008 & 0.353 $\pm$ 0.005 & 0.338 $\pm$ 0.023 & 0.490 & 0.279 $\pm$ 0.006 & 0.645 $\pm$ 0.002 & 0.138 $\pm$ 0.003 & 0.137 $\pm$ 0.001 & 0.620 & 0.085 $\pm$ 0.001 \\
    \cmidrule(lr){1-11}
    \multicolumn{11}{l}{\textit{Sequence-based [1h]}} \\
    \cmidrule(lr){1-11}
    GRU & 0.780 $\pm$ 0.008 & 0.474 $\pm$ 0.014 & 0.473 $\pm$ 0.008 & 0.586 & 0.417 $\pm$ 0.006 & 0.855 $\pm$ 0.005 & 0.246 $\pm$ 0.007 & 0.363 $\pm$ 0.025 & 0.821 & 0.359 $\pm$ 0.011 \\
    LSTM & 0.779 $\pm$ 0.009 & 0.470 $\pm$ 0.004 & 0.480 $\pm$ 0.001 & 0.583 & 0.421 $\pm$ 0.010 & 0.842 $\pm$ 0.005 & 0.241 $\pm$ 0.019 & 0.355 $\pm$ 0.004 & 0.823 & 0.354 $\pm$ 0.008 \\
    TCN & 0.780 $\pm$ 0.014 & 0.473 $\pm$ 0.007 & 0.470 $\pm$ 0.013 & 0.557 & 0.421 $\pm$ 0.015 & 0.855 $\pm$ 0.004 & 0.263 $\pm$ 0.008 & 0.366 $\pm$ 0.015 & 0.810 & 0.347 $\pm$ 0.012 \\
    Transformer & 0.767 $\pm$ 0.005 & 0.464 $\pm$ 0.004 & 0.454 $\pm$ 0.008 & 0.573 & 0.391 $\pm$ 0.013 & 0.806 $\pm$ 0.001 & 0.230 $\pm$ 0.018 & 0.300 $\pm$ 0.007 & 0.744 & 0.260 $\pm$ 0.015 \\
    SSM & 0.741 $\pm$ 0.002 & 0.449 $\pm$ 0.009 & 0.451 $\pm$ 0.008 & 0.519 & 0.356 $\pm$ 0.005 & 0.788 $\pm$ 0.003 & 0.212 $\pm$ 0.011 & 0.289 $\pm$ 0.005 & 0.796 & 0.233 $\pm$ 0.007 \\
    \cmidrule(lr){1-11}
    \multicolumn{11}{l}{\textit{Graph [\textasciitilde{}1h]}} \\
    \cmidrule(lr){1-11}
    RAINDROP & 0.725 $\pm$ 0.010 & 0.434 $\pm$ 0.013 & 0.420 $\pm$ 0.013 & 0.567 & 0.332 $\pm$ 0.009 & 0.752 $\pm$ 0.007 & 0.228 $\pm$ 0.007 & 0.219 $\pm$ 0.009 & 0.683 & 0.151 $\pm$ 0.010 \\
    \bottomrule
\end{tabular}
}
\end{table}

\clearpage
\subsection{Calibration plots \emph{MELODY-KISPI}}
\label{app:calibration_Private}

\begin{figure}[h!]
    \centering
    \includegraphics[width=1\linewidth]{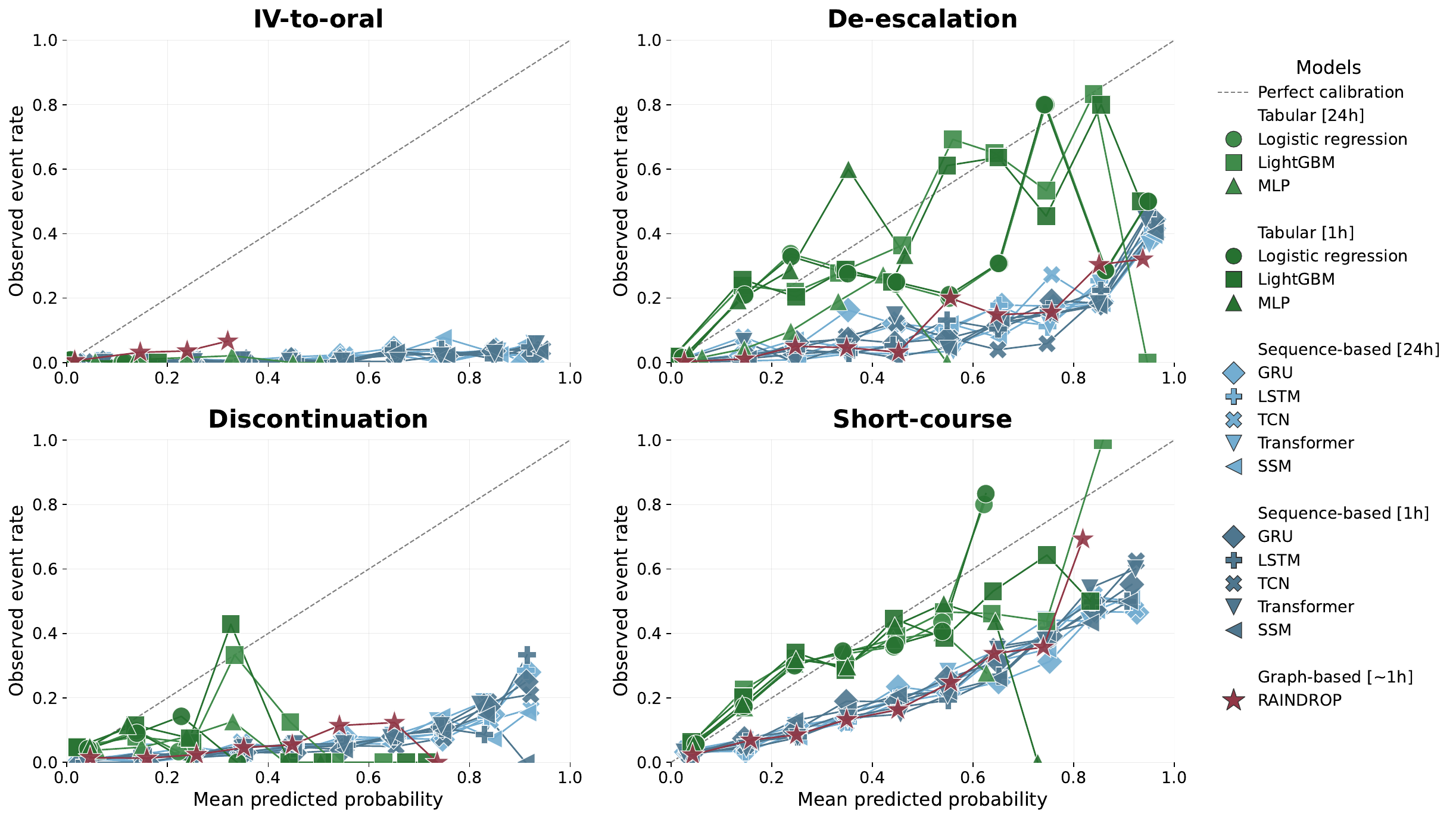}
    \caption{Calibration plots for tabular, sequence-based, and graph-based temporal models across temporal representations for the four AMS intervention predictions tasks in the \emph{MELODY-KISPI} cohort.}
    \label{fig:calibration_private}
\end{figure}

\subsection{Multi-task learning for AMS intervention prediction}
\label{app:mtl}
\begin{figure}[H]
    \centering
    \includegraphics[width=1\linewidth]{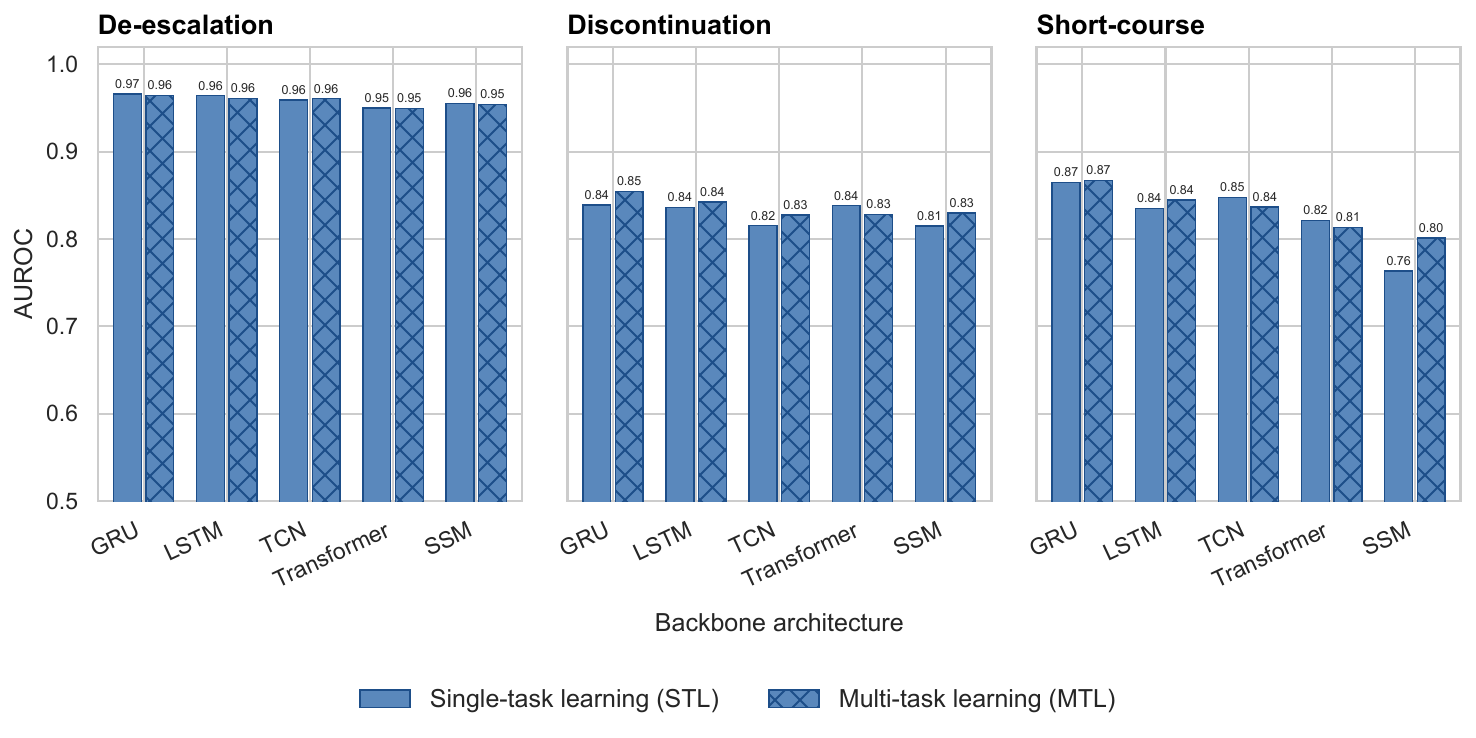}
    \caption{Comparison of single-task and multi-task learning for sequence models across three AMS targets on the \emph{PIC} cohort.}
    \label{fig:auroc_stl_mtl_PIC}
\end{figure}

\begin{figure}[H]
    \centering
    \includegraphics[width=1\linewidth]{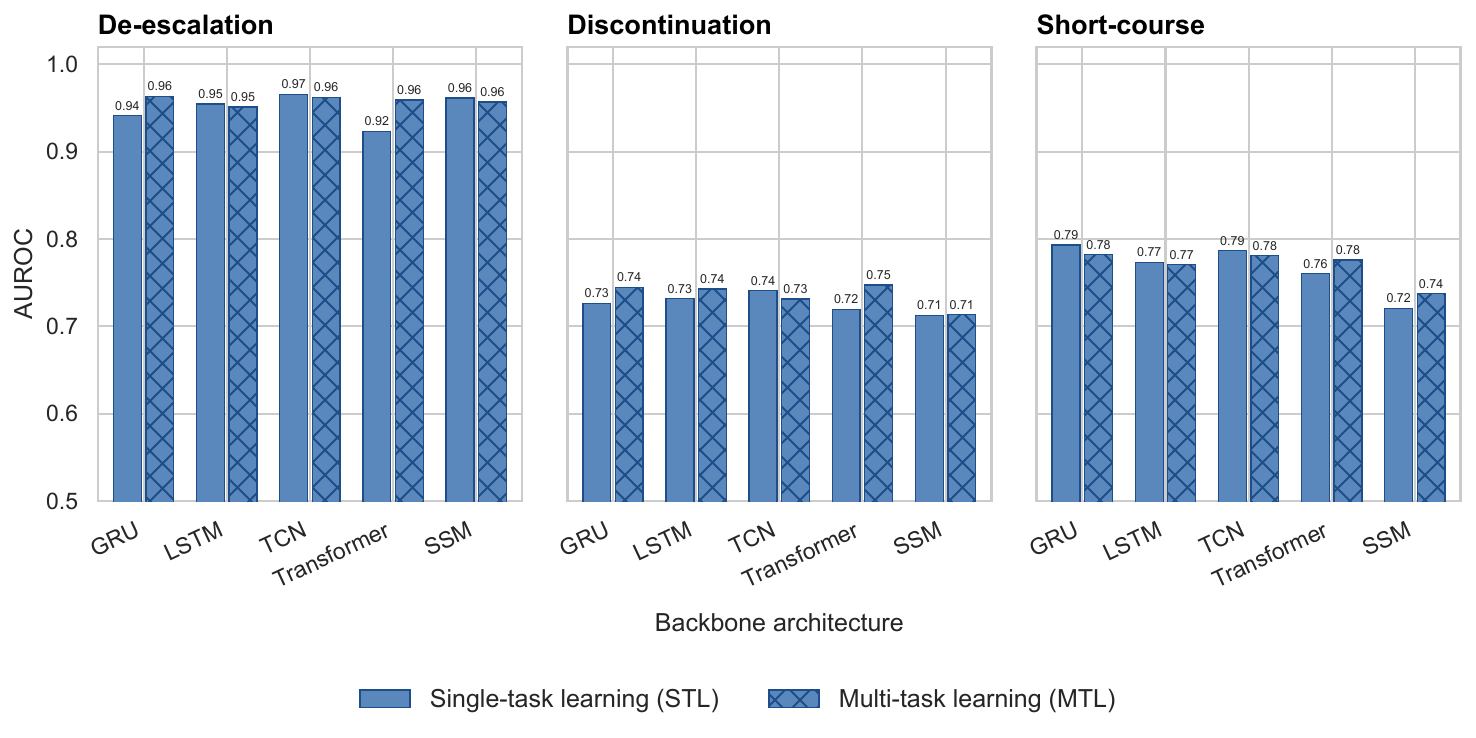}
    \caption{Comparison of single-task and multi-task learning for sequence models across three AMS targets on the \emph{MELODY-KISPI} cohort.}
    \label{fig:auroc_stl_mtl_private}
\end{figure}

We finally evaluate whether multi-task learning can improve performance by jointly modeling multiple stewardship targets. Since these interventions are clinically related and sometimes considered jointly (e.g., intervention vs no intervention), we hypothesize that sharing representations across tasks may improve learning. We therefore compare STL and MTL on targets de-escalation, discontinuation, and short-course as these co-occur (Figure~\ref{fig:UpSet_overlap_outcomes}) as we expect them to be more related to each other than to IV-to-oral. We use sequence models using 24-hour representations. AUROC performance across cohorts and training both settings are shown in Figures \ref{fig:auroc_stl_mtl_PIC} and \ref{fig:auroc_stl_mtl_private}. Full results are described in section \ref{app:mtl_detailed_results}.

\clearpage
\subsubsection{MTL modeling detailed results}
\label{app:mtl_detailed_results}
\begin{table}[h!]
    \centering
    \scriptsize
    \caption{MTL results for de-escalation.}
    \label{tab:MTL_detailed_deescalation_results}
  \resizebox{0.7\textwidth}{!}{
\begin{tabular}{ll*{6}{c}}
    \toprule
    & & \multicolumn{6}{c}{De-escalation} \\
    \cmidrule(lr){3-8}
    & & \multicolumn{3}{c}{MELODY-KISPI} & \multicolumn{3}{c}{PIC} \\
    \cmidrule(lr){3-5}\cmidrule(lr){6-8}
    \textbf{Architecture} & \textbf{Approach} & \textbf{AUROC} & \textbf{F1} & \textbf{AUPRC} & \textbf{AUROC} & \textbf{F1} & \textbf{AUPRC} \\
    \midrule
    \multicolumn{8}{l}{\quad\textit{Sequential [24h]}} \\
    GRU & hard-sharing & 0.955 & 0.464 & 0.520 & 0.964 & 0.433 & 0.549 \\
    GRU & uncertainty-weighted & 0.956 & 0.469 & 0.525 & 0.965 & 0.447 & 0.550 \\
    GRU & MMoE+PCGrad & 0.962 & 0.458 & \textbf{0.595} & 0.965 & 0.443 & 0.560 \\
    LSTM & hard-sharing & 0.951 & 0.447 & 0.540 & 0.961 & 0.404 & 0.511 \\
    LSTM & uncertainty-weighted & 0.949 & 0.423 & 0.516 & 0.962 & 0.403 & 0.511 \\
    LSTM & MMoE+PCGrad & 0.958 & 0.456 & 0.558 & 0.964 & 0.410 & 0.537 \\
    TCN & hard-sharing & 0.962 & 0.499 & 0.525 & 0.961 & 0.395 & 0.512 \\
    TCN & uncertainty-weighted & 0.959 & 0.511 & 0.510 & 0.962 & 0.434 & 0.536 \\
    TCN & MMoE+PCGrad & 0.960 & 0.482 & 0.515 & 0.964 & 0.424 & 0.538 \\
    Transformer & hard-sharing & 0.959 & 0.465 & 0.548 & 0.949 & 0.331 & 0.397 \\
    Transformer & uncertainty-weighted & 0.955 & 0.463 & 0.507 & 0.951 & 0.351 & 0.432 \\
    Transformer & MMoE+PCGrad & 0.956 & 0.423 & 0.545 & 0.948 & 0.349 & 0.389 \\
    SSM & hard-sharing & 0.957 & 0.469 & 0.511 & 0.954 & 0.352 & 0.443 \\
    SSM & uncertainty-weighted & 0.958 & 0.477 & 0.536 & 0.951 & 0.328 & 0.416 \\
    SSM & MMoE+PCGrad & 0.953 & 0.465 & 0.493 & 0.943 & 0.304 & 0.391 \\
    \bottomrule
\end{tabular}
}
\end{table}

\begin{table}[h!]
    \centering
    \scriptsize
    \caption{MTL results for discontinuation.}
    \label{tab:MTL_detailed_discontinuation_results}
  \resizebox{0.7\textwidth}{!}{
\begin{tabular}{ll*{6}{c}}
    \toprule
    & & \multicolumn{6}{c}{Discontinuation} \\
    \cmidrule(lr){3-8}
    & & \multicolumn{3}{c}{MELODY-KISPI} & \multicolumn{3}{c}{PIC} \\
    \cmidrule(lr){3-5}\cmidrule(lr){6-8}
    \textbf{Architecture} & \textbf{Approach} & \textbf{AUROC} & \textbf{F1} & \textbf{AUPRC} & \textbf{AUROC} & \textbf{F1} & \textbf{AUPRC} \\
    \midrule
    \multicolumn{8}{l}{\quad\textit{Sequential [24h]}} \\
    GRU & hard-sharing & 0.742 & 0.234 & 0.234 & 0.855 & 0.099 & 0.106 \\
    GRU & uncertainty-weighted & 0.742 & 0.233 & 0.237 & 0.855 & 0.102 & 0.104 \\
    GRU & MMoE+PCGrad & 0.743 & 0.229 & 0.212 & 0.860 & 0.102 & 0.115 \\
    LSTM & hard-sharing & 0.743 & \textbf{0.248} & 0.199 & 0.843 & 0.106 & 0.100 \\
    LSTM & uncertainty-weighted & 0.730 & 0.226 & 0.187 & 0.844 & 0.105 & 0.100 \\
    LSTM & MMoE+PCGrad & 0.743 & 0.243 & 0.180 & 0.854 & 0.105 & 0.121 \\
    TCN & hard-sharing & 0.731 & 0.238 & \textbf{0.228} & 0.828 & 0.091 & 0.075 \\
    TCN & uncertainty-weighted & 0.743 & 0.248 & 0.209 & 0.841 & 0.096 & 0.095 \\
    TCN & MMoE+PCGrad & 0.740 & 0.226 & 0.188 & 0.849 & 0.092 & 0.101 \\
    Transformer & hard-sharing & 0.747 & 0.239 & 0.208 & 0.828 & 0.096 & 0.092 \\
    Transformer & uncertainty-weighted & 0.732 & 0.242 & 0.201 & 0.844 & 0.086 & 0.111 \\
    Transformer & MMoE+PCGrad & 0.722 & 0.226 & 0.208 & 0.822 & 0.083 & 0.081 \\
    SSM & hard-sharing & 0.713 & 0.238 & 0.173 & 0.830 & 0.101 & 0.092 \\
    SSM & uncertainty-weighted & 0.717 & 0.246 & 0.177 & 0.815 & 0.094 & 0.086 \\
    SSM & MMoE+PCGrad & 0.699 & 0.217 & 0.164 & 0.824 & 0.087 & 0.093 \\
    \bottomrule
\end{tabular}
}
\end{table}

\begin{table}[h!]
    \centering
    \scriptsize
    \caption{MTL results for short-courses.}
    \label{tab:MTL_detailed_shortcourse_results}
  \resizebox{0.7\textwidth}{!}{
\begin{tabular}{ll*{6}{c}}
    \toprule
    & & \multicolumn{6}{c}{Short-course} \\
    \cmidrule(lr){3-8}
    & & \multicolumn{3}{c}{MELODY-KISPI} & \multicolumn{3}{c}{PIC} \\
    \cmidrule(lr){3-5}\cmidrule(lr){6-8}
    \textbf{Architecture} & \textbf{Approach} & \textbf{AUROC} & \textbf{F1} & \textbf{AUPRC} & \textbf{AUROC} & \textbf{F1} & \textbf{AUPRC} \\
    \midrule
    \multicolumn{8}{l}{\quad\textit{Sequential [24h]}} \\
    GRU & hard-sharing & 0.774 & 0.501 & 0.457 & 0.866 & 0.287 & 0.423 \\
    GRU & uncertainty-weighted & 0.775 & 0.504 & 0.459 & 0.865 & 0.285 & 0.422 \\
    GRU & MMoE+PCGrad & 0.779 & 0.501 & \textbf{0.474} & 0.867 & 0.280 & 0.427 \\
    LSTM & hard-sharing & 0.771 & 0.497 & 0.454 & 0.845 & \textbf{0.274} & 0.367 \\
    LSTM & uncertainty-weighted & 0.767 & 0.482 & 0.458 & 0.845 & 0.271 & 0.366 \\
    LSTM & MMoE+PCGrad & 0.780 & 0.511 & 0.458 & 0.864 & 0.260 & 0.402 \\
    TCN & hard-sharing & 0.781 & \textbf{0.519} & 0.460 & 0.837 & 0.264 & 0.322 \\
    TCN & uncertainty-weighted & 0.776 & 0.516 & 0.450 & 0.848 & 0.267 & 0.343 \\
    TCN & MMoE+PCGrad & 0.768 & 0.497 & 0.459 & 0.855 & 0.254 & 0.360 \\
    Transformer & hard-sharing & 0.776 & 0.496 & 0.451 & 0.813 & 0.229 & 0.263 \\
    Transformer & uncertainty-weighted & 0.763 & 0.494 & 0.456 & 0.827 & 0.223 & 0.305 \\
    Transformer & MMoE+PCGrad & 0.762 & 0.491 & 0.446 & 0.810 & 0.227 & 0.253 \\
    SSM & hard-sharing & 0.737 & 0.481 & 0.417 & 0.801 & 0.235 & 0.282 \\
    SSM & uncertainty-weighted & 0.739 & 0.482 & 0.421 & 0.790 & 0.227 & 0.248 \\
    SSM & MMoE+PCGrad & 0.724 & 0.465 & 0.393 & 0.788 & 0.232 & 0.249 \\
    \bottomrule
\end{tabular}
}
\end{table}

\clearpage
\subsection{When do sequence models help}
\label{app:when_sequence_models_help}

\begin{figure}[ht]
    \centering
    \includegraphics[width=0.8\linewidth]{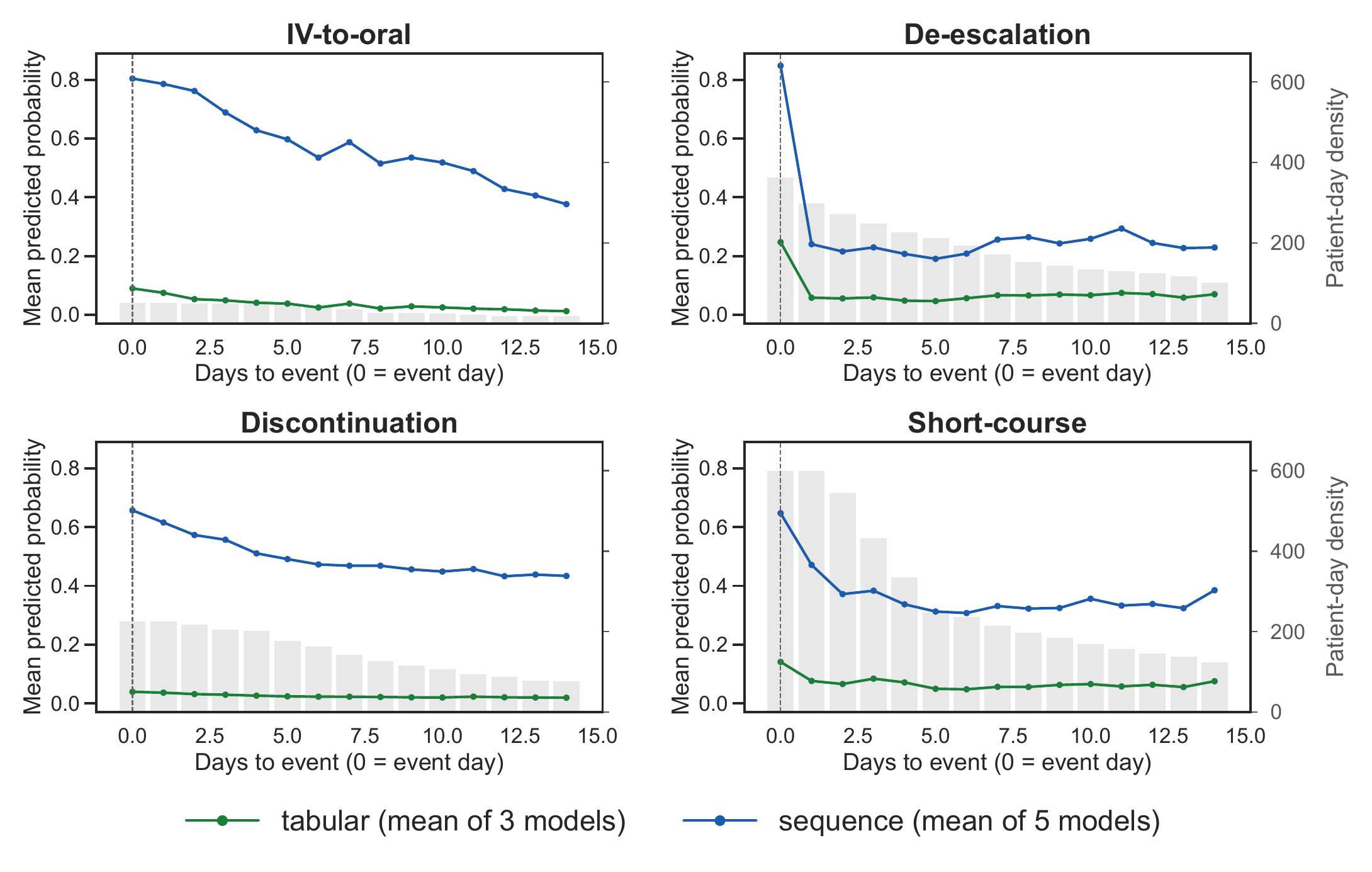}
    \caption{Average event aligned probabilities of tabular and sequence models of \emph{PIC} targets. Tabular models include logistic regression, LightGBM, and the MLP. Sequence models include the GRU, LSTM, TCN, Transformer, and SSM. Grey bars behind show the density of as the number of patient-days.}
    \label{fig:event_aligned_probabilities_pic}
\end{figure}

\begin{figure}[ht]
    \centering
    \includegraphics[width=0.8\linewidth]{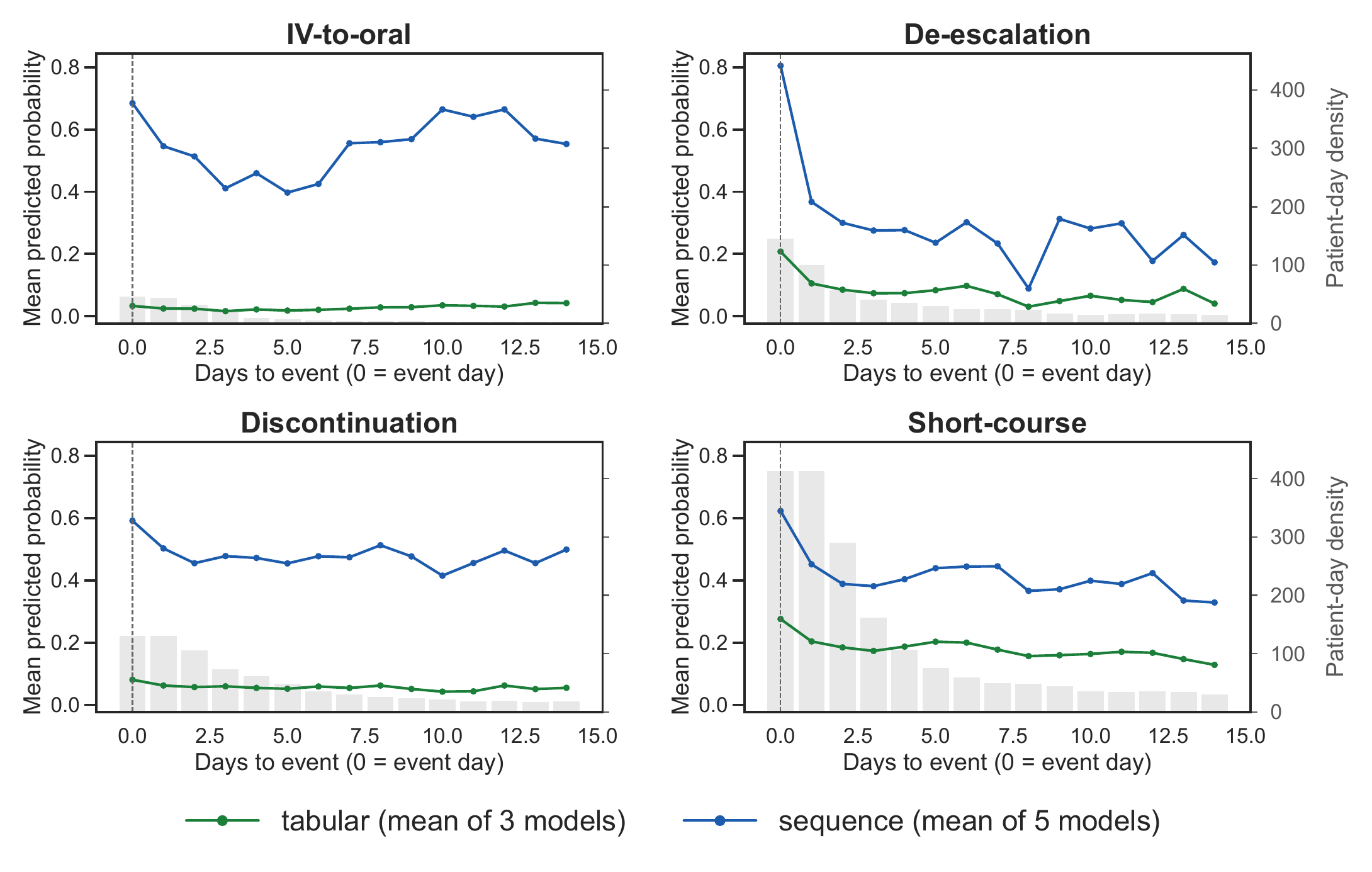}
    \caption{Average event aligned probabilities of tabular and sequence models of \emph{MELODY-KISPI} targets.}
    \label{fig:event_aligned_probabilities_private}
\end{figure}

\begin{figure}[!ht]
    \centering
    \includegraphics[width=1\linewidth]{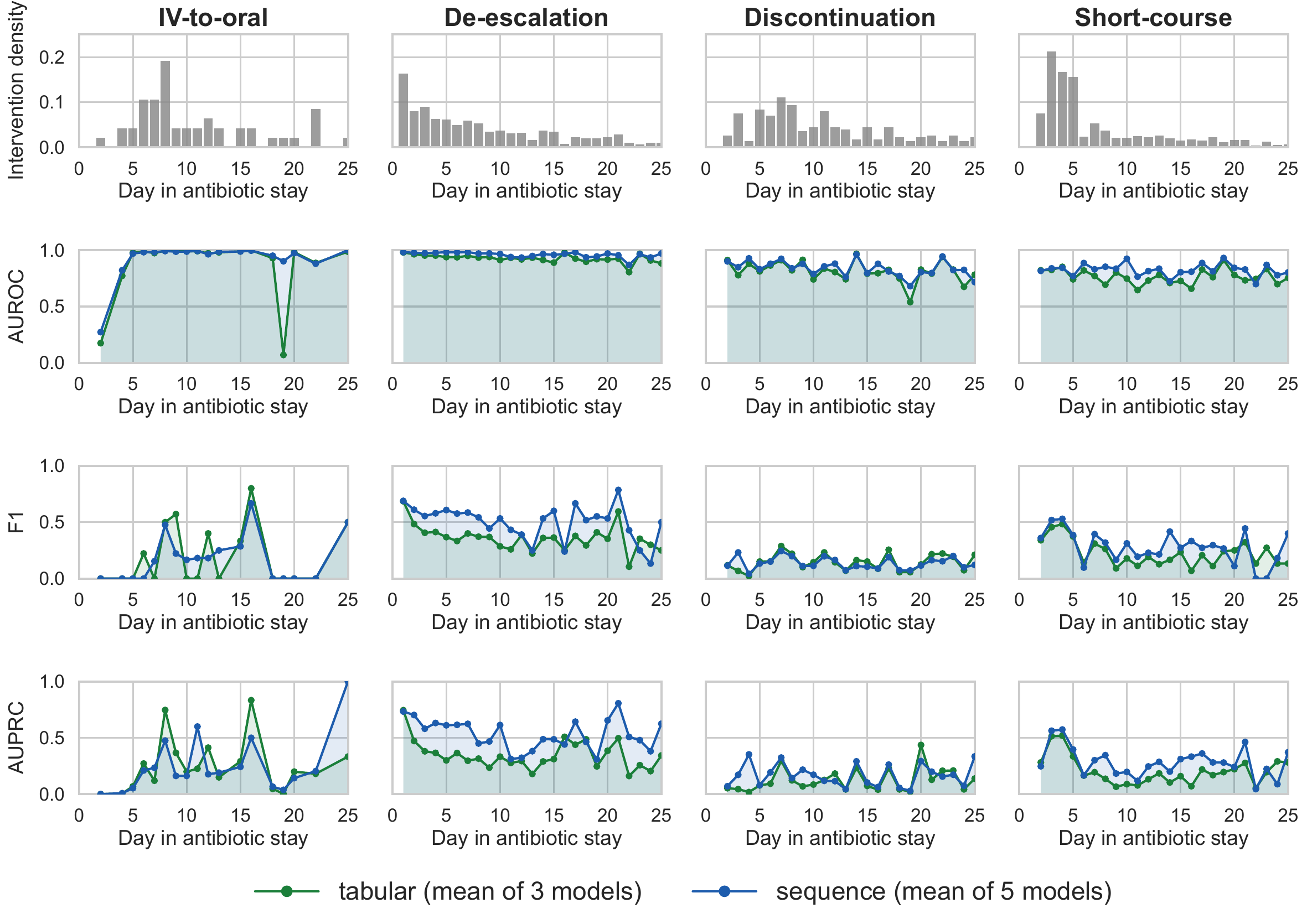}
    \caption{Average prediction performance of tabular and sequence model groups across antibiotic days in a stay in \emph{PIC}. Tabular models include Logistic regression, LightGBM, and the MLP. Sequence models include the GRU, LSTM, TCN, Transformer, and SSM.}
    \label{fig:intervention_and_performance_by_day_pic}
\end{figure}

\begin{figure} [!ht]
    \centering
    \includegraphics[width=1\linewidth]{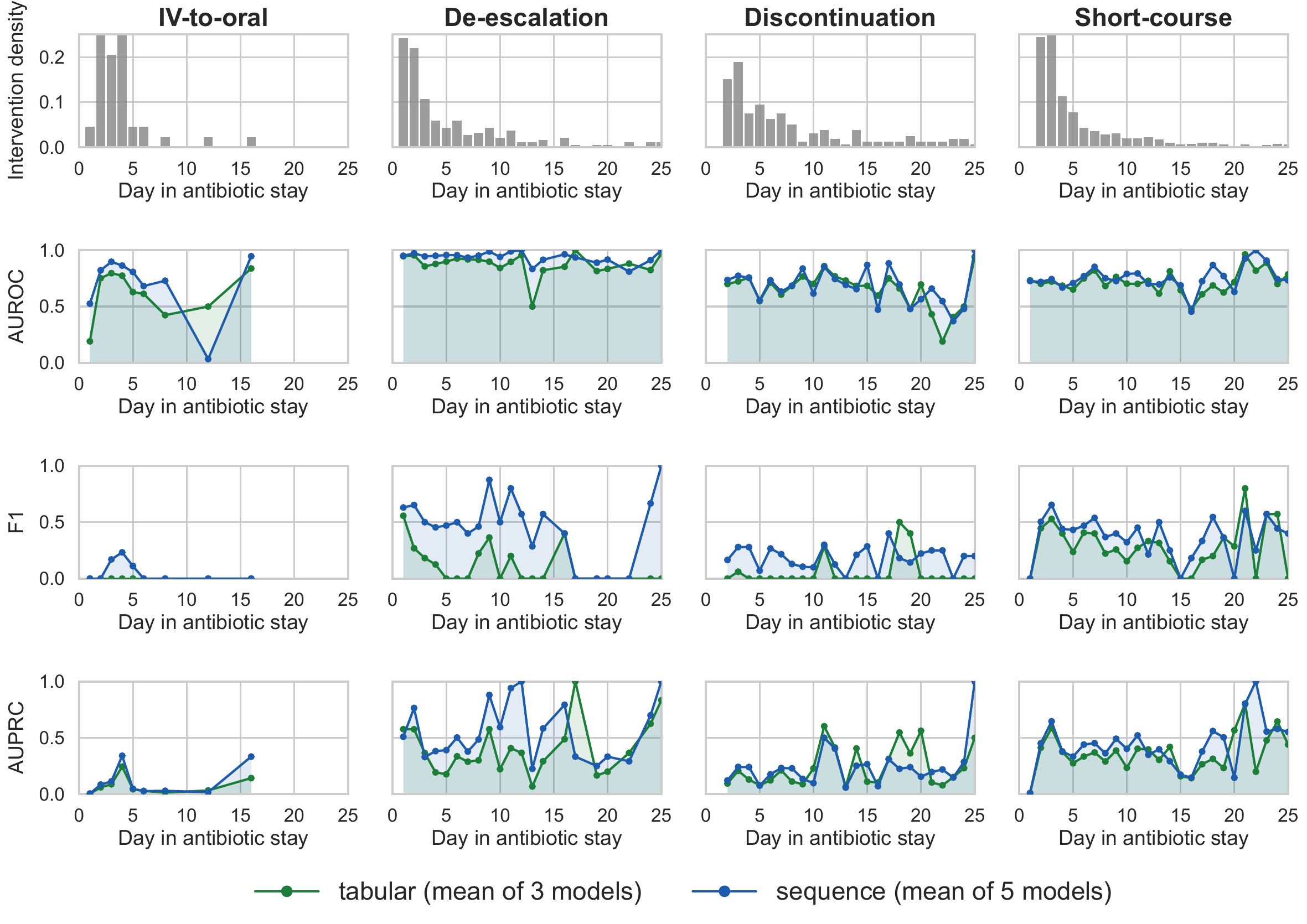}
    \caption{Average prediction performance of tabular and sequence model groups across antibiotic days in a stay in \emph{MELODY-KISPI}. Tabular models include Logistic regression, LightGBM, and the MLP. Sequence models include the GRU, LSTM, TCN, Transformer, and SSM.}
    \label{fig:intervention_and_performance_by_day_private}
\end{figure}

\clearpage
\subsection{1h encoder ablations}
\label{app:1h ablations}

To better understand the added benefit of finer-grained 1h data and the learnable event grid encoder we performed a set of ablations (Figures \ref{fig:1h_ablations_pic} and \ref{fig:1h_ablations_private} below). We here compare the daily sequence-based [24h] model family on daily aggregated data, adding the 1h CNN learnable encoder such as described above in Section~\ref{app:impl_details_1h_bins}, and compare to the following two ablations;

\paragraph{1h linear projection encoder.}
This encoder keeps the day-level sequence architecture used by the CNN longitudinal model and changes only how each day’s hourly event grid is compressed into an embedding. We use the default hourly chart/lab measurement set with LOCF, binary indicators of whether a new observation occurred in that bin, and the hours since the most recent observation. The resulting high-dimensional tensor is flattened through a linear projection into the 64-dimensional embedding that is concatenated to the day's tabular features.

\paragraph{1h direct encoder.}
This encoder does not use a day-level event-grid encoder. Instead it models the recent history as an hourly sequence and fuses tabular features only after the sequence trunk (late fusion). Starting from the same per-day 24-bin grids (value / observation mask / hours-since-last-observation planes, we stack up to 5 consecutive prior days and reshape them into a single sequence. All grid planes and tabular features are standardized on the training split. A sequence backbone (same five architectures as above; hidden size 128, 2 layers, dropout 0.2) reads the raw hourly feature vectors directly. The hidden state at the last valid hour is concatenated with the current day’s tabular vector and mapped to a single patient-day logit. Relative to the longitudinal CNN/linear models, the direct hourly modeling differs in three structural ways: (1) the sequence unit is an hour, not a day; (2) multi-day context is a fixed 5-day hourly stack, not a day-level sequence over the admission; (3) there is no learned day-level grid encoder, tabular information enters only after temporal encoding.

\begin{figure}[!h]
    \centering
    \includegraphics[width=0.85\linewidth]{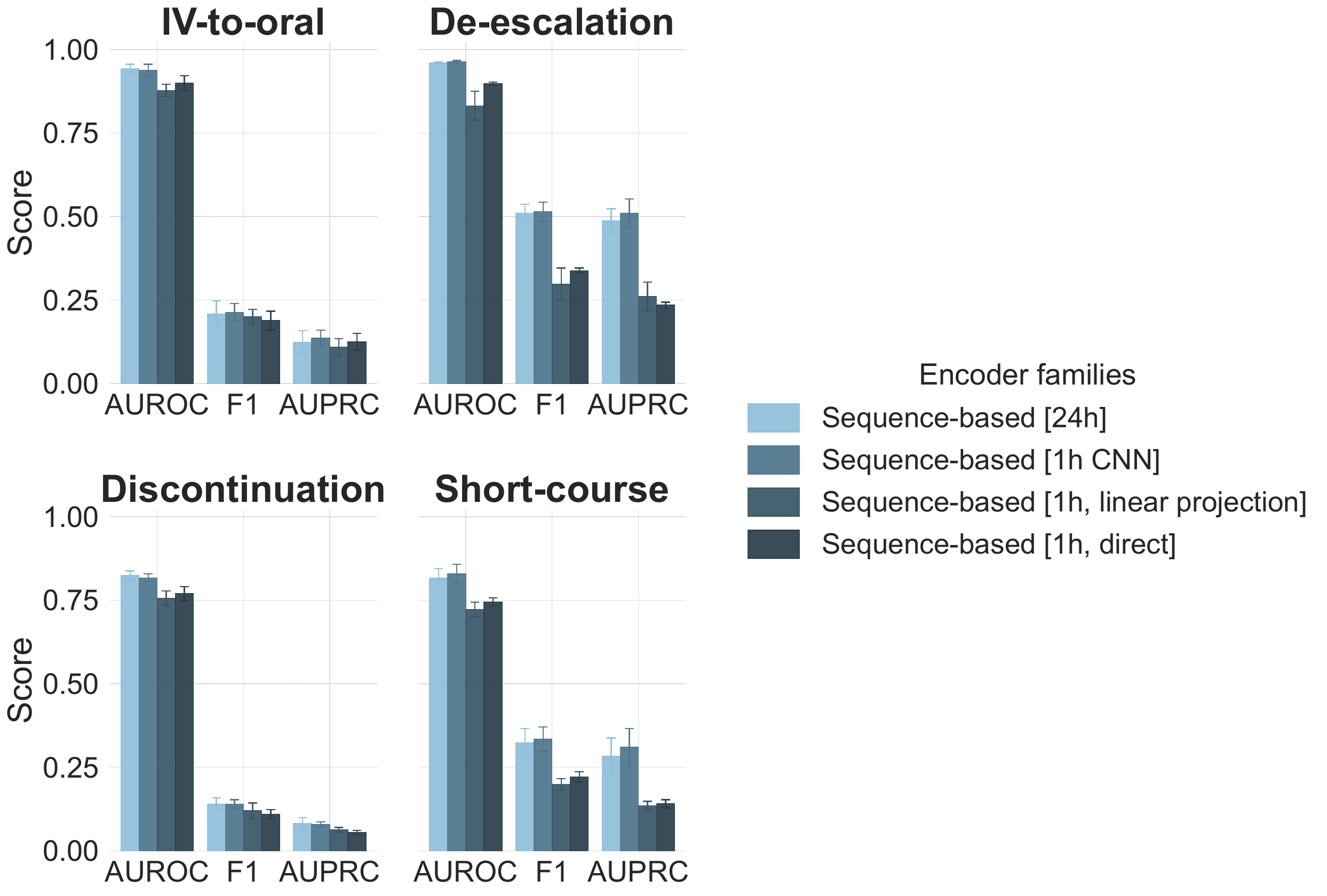}
    \caption{Encoder ablations of 1h sequence models on \emph{PIC} data. Bars represent the average performance metrics and error bars show standard deviation across models and seeds.}
    \label{fig:1h_ablations_pic}
\end{figure}

\begin{figure}[!h]
    \centering
    \includegraphics[width=0.85\linewidth]{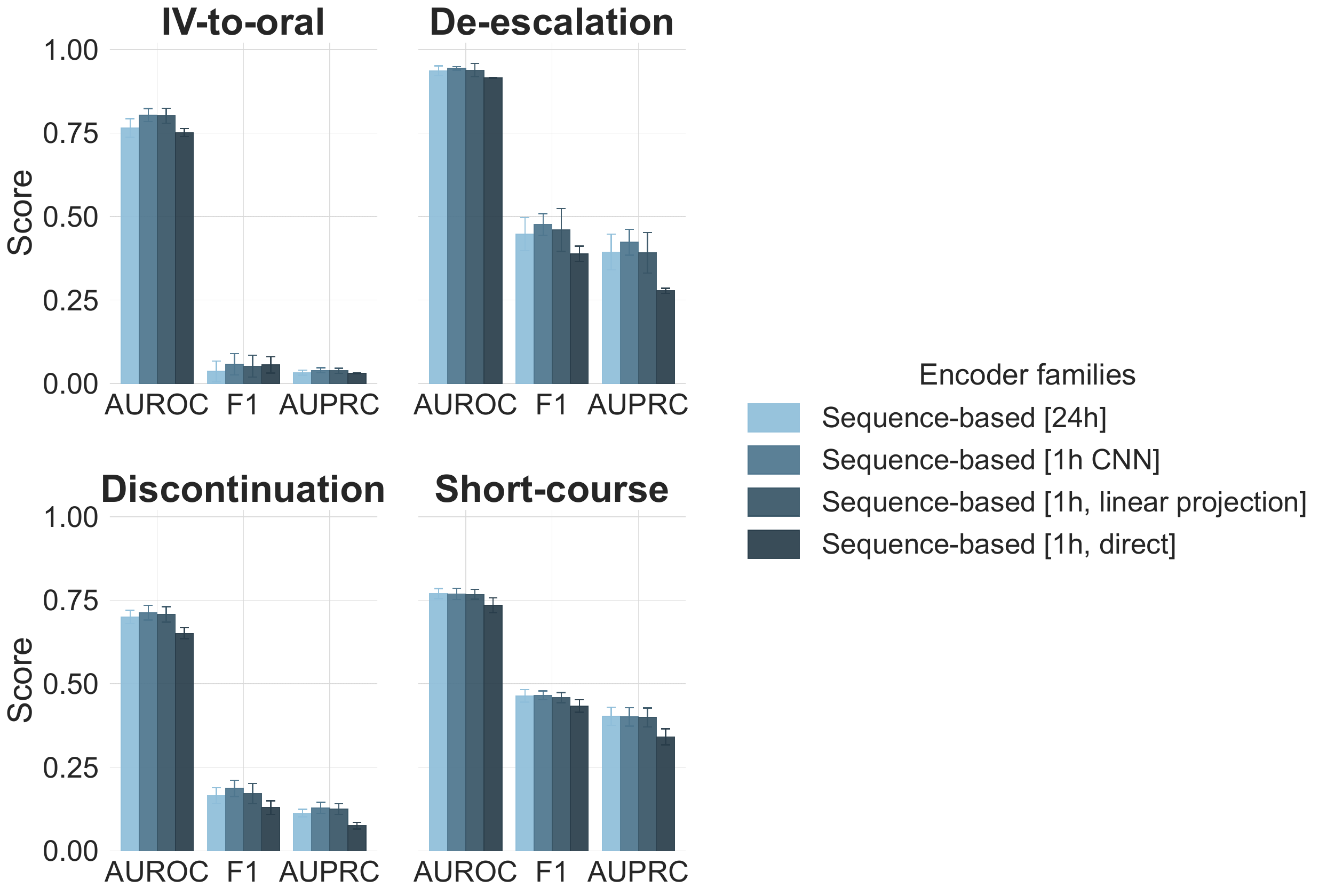}
    \caption{Encoder ablations of 1h sequence models on \emph{MELODY-KISPI} data. Bars represent the average performance metrics and error bars show standard deviation across models and seeds.}
    \label{fig:1h_ablations_private}
\end{figure}

\clearpage
\subsection{Patient-temporal splitting}
\label{app: patient-temporal_splitting}

\begin{table}[h!]
    \centering
    \scriptsize
    \caption{Comparing patient-random and patient-temporal splitting in \emph{MELODY-KISPI}. For patient-temporal splitting, patients are ordered by admission time, and a hard cutoff time \textit{T} is set at the last 20\% of patients. Train $< T$, test $\geq T$. AUROC, F1 score (threshold-optimized), and AUPRC are reported with mean scores and standard deviation across three seeds.}
    \label{tab:split_strategy_private}
  \resizebox{\textwidth}{!}{\begin{tabular}{llcccccc}
\toprule
Target & Model & \multicolumn{2}{c}{AUROC} & \multicolumn{2}{c}{F1} & \multicolumn{2}{c}{AUPRC} \\
\cmidrule(lr){3-4} \cmidrule(lr){5-6} \cmidrule(lr){7-8}
& & random & temporal & random & temporal & random & temporal \\
\midrule
IV-to-oral & Logistic regression & 0.847 $\pm$ 0.000 & 0.780 $\pm$ 0.000 & 0.083 $\pm$ 0.000 & 0.032 $\pm$ 0.000 & 0.065 $\pm$ 0.000 & 0.034 $\pm$ 0.000 \\
IV-to-oral & LightGBM & 0.797 $\pm$ 0.000 & 0.756 $\pm$ 0.000 & 0.000 $\pm$ 0.000 & 0.000 $\pm$ 0.000 & 0.097 $\pm$ 0.000 & 0.035 $\pm$ 0.000 \\
IV-to-oral & MLP & 0.547 $\pm$ 0.070 & 0.583 $\pm$ 0.047 & 0.025 $\pm$ 0.009 & 0.019 $\pm$ 0.006 & 0.014 $\pm$ 0.002 & 0.010 $\pm$ 0.002 \\
IV-to-oral & GRU & 0.836 $\pm$ 0.013 & 0.753 $\pm$ 0.018 & 0.090 $\pm$ 0.014 & 0.033 $\pm$ 0.031 & 0.064 $\pm$ 0.008 & 0.026 $\pm$ 0.001 \\
IV-to-oral & LSTM & 0.838 $\pm$ 0.008 & 0.743 $\pm$ 0.001 & 0.096 $\pm$ 0.015 & 0.031 $\pm$ 0.027 & 0.064 $\pm$ 0.015 & 0.027 $\pm$ 0.001 \\
IV-to-oral & TCN & 0.836 $\pm$ 0.008 & 0.739 $\pm$ 0.013 & 0.115 $\pm$ 0.004 & 0.000 $\pm$ 0.000 & 0.070 $\pm$ 0.006 & 0.028 $\pm$ 0.002 \\
IV-to-oral & Transformer & 0.787 $\pm$ 0.059 & 0.781 $\pm$ 0.013 & 0.069 $\pm$ 0.034 & 0.039 $\pm$ 0.040 & 0.058 $\pm$ 0.017 & 0.039 $\pm$ 0.009 \\
IV-to-oral & SSM & 0.831 $\pm$ 0.005 & 0.791 $\pm$ 0.014 & 0.100 $\pm$ 0.042 & 0.041 $\pm$ 0.039 & 0.079 $\pm$ 0.013 & 0.038 $\pm$ 0.004 \\
\addlinespace
De-escalation & Logistic regression & 0.900 $\pm$ 0.000 & 0.903 $\pm$ 0.000 & 0.388 $\pm$ 0.000 & 0.377 $\pm$ 0.000 & 0.282 $\pm$ 0.000 & 0.274 $\pm$ 0.000 \\
De-escalation & LightGBM & 0.932 $\pm$ 0.000 & 0.945 $\pm$ 0.000 & 0.241 $\pm$ 0.000 & 0.315 $\pm$ 0.000 & 0.432 $\pm$ 0.000 & 0.415 $\pm$ 0.000 \\
De-escalation & MLP & 0.723 $\pm$ 0.064 & 0.812 $\pm$ 0.090 & 0.255 $\pm$ 0.070 & 0.308 $\pm$ 0.088 & 0.172 $\pm$ 0.054 & 0.226 $\pm$ 0.073 \\
De-escalation & GRU & 0.943 $\pm$ 0.002 & 0.946 $\pm$ 0.000 & 0.491 $\pm$ 0.018 & 0.472 $\pm$ 0.013 & 0.413 $\pm$ 0.019 & 0.412 $\pm$ 0.010 \\
De-escalation & LSTM & 0.937 $\pm$ 0.004 & 0.943 $\pm$ 0.006 & 0.485 $\pm$ 0.014 & 0.483 $\pm$ 0.022 & 0.411 $\pm$ 0.024 & 0.410 $\pm$ 0.029 \\
De-escalation & TCN & 0.937 $\pm$ 0.003 & 0.945 $\pm$ 0.005 & 0.454 $\pm$ 0.024 & 0.483 $\pm$ 0.012 & 0.405 $\pm$ 0.008 & 0.398 $\pm$ 0.029 \\
De-escalation & Transformer & 0.933 $\pm$ 0.005 & 0.938 $\pm$ 0.011 & 0.418 $\pm$ 0.035 & 0.440 $\pm$ 0.035 & 0.386 $\pm$ 0.015 & 0.405 $\pm$ 0.035 \\
De-escalation & SSM & 0.932 $\pm$ 0.002 & 0.936 $\pm$ 0.006 & 0.400 $\pm$ 0.021 & 0.437 $\pm$ 0.035 & 0.378 $\pm$ 0.031 & 0.427 $\pm$ 0.040 \\
\addlinespace
Discontinuation & Logistic regression & 0.665 $\pm$ 0.000 & 0.654 $\pm$ 0.000 & 0.139 $\pm$ 0.000 & 0.126 $\pm$ 0.000 & 0.074 $\pm$ 0.000 & 0.080 $\pm$ 0.000 \\
Discontinuation & LightGBM & 0.670 $\pm$ 0.000 & 0.665 $\pm$ 0.000 & 0.084 $\pm$ 0.000 & 0.065 $\pm$ 0.000 & 0.085 $\pm$ 0.000 & 0.090 $\pm$ 0.000 \\
Discontinuation & MLP & 0.623 $\pm$ 0.007 & 0.603 $\pm$ 0.013 & 0.121 $\pm$ 0.004 & 0.126 $\pm$ 0.015 & 0.073 $\pm$ 0.002 & 0.074 $\pm$ 0.006 \\
Discontinuation & GRU & 0.721 $\pm$ 0.024 & 0.739 $\pm$ 0.007 & 0.170 $\pm$ 0.022 & 0.199 $\pm$ 0.017 & 0.113 $\pm$ 0.025 & 0.139 $\pm$ 0.003 \\
Discontinuation & LSTM & 0.707 $\pm$ 0.019 & 0.711 $\pm$ 0.007 & 0.151 $\pm$ 0.018 & 0.173 $\pm$ 0.012 & 0.100 $\pm$ 0.007 & 0.126 $\pm$ 0.006 \\
Discontinuation & TCN & 0.714 $\pm$ 0.014 & 0.717 $\pm$ 0.013 & 0.168 $\pm$ 0.017 & 0.180 $\pm$ 0.025 & 0.119 $\pm$ 0.002 & 0.126 $\pm$ 0.003 \\
Discontinuation & Transformer & 0.702 $\pm$ 0.010 & 0.703 $\pm$ 0.006 & 0.153 $\pm$ 0.012 & 0.165 $\pm$ 0.019 & 0.101 $\pm$ 0.007 & 0.115 $\pm$ 0.003 \\
Discontinuation & SSM & 0.711 $\pm$ 0.023 & 0.676 $\pm$ 0.013 & 0.177 $\pm$ 0.022 & 0.165 $\pm$ 0.015 & 0.110 $\pm$ 0.019 & 0.113 $\pm$ 0.010 \\
\addlinespace
Short-course & Logistic regression & 0.732 $\pm$ 0.000 & 0.726 $\pm$ 0.000 & 0.406 $\pm$ 0.000 & 0.427 $\pm$ 0.000 & 0.335 $\pm$ 0.000 & 0.338 $\pm$ 0.000 \\
Short-course & LightGBM & 0.762 $\pm$ 0.000 & 0.754 $\pm$ 0.000 & 0.312 $\pm$ 0.000 & 0.327 $\pm$ 0.000 & 0.364 $\pm$ 0.000 & 0.383 $\pm$ 0.000 \\
Short-course & MLP & 0.712 $\pm$ 0.036 & 0.707 $\pm$ 0.044 & 0.406 $\pm$ 0.030 & 0.400 $\pm$ 0.037 & 0.321 $\pm$ 0.028 & 0.314 $\pm$ 0.035 \\
Short-course & GRU & 0.791 $\pm$ 0.007 & 0.773 $\pm$ 0.005 & 0.463 $\pm$ 0.011 & 0.467 $\pm$ 0.018 & 0.419 $\pm$ 0.007 & 0.405 $\pm$ 0.006 \\
Short-course & LSTM & 0.796 $\pm$ 0.003 & 0.772 $\pm$ 0.008 & 0.485 $\pm$ 0.009 & 0.475 $\pm$ 0.011 & 0.422 $\pm$ 0.001 & 0.403 $\pm$ 0.010 \\
Short-course & TCN & 0.799 $\pm$ 0.003 & 0.766 $\pm$ 0.017 & 0.485 $\pm$ 0.007 & 0.458 $\pm$ 0.024 & 0.438 $\pm$ 0.012 & 0.405 $\pm$ 0.025 \\
Short-course & Transformer & 0.778 $\pm$ 0.006 & 0.756 $\pm$ 0.015 & 0.451 $\pm$ 0.016 & 0.454 $\pm$ 0.019 & 0.395 $\pm$ 0.007 & 0.370 $\pm$ 0.019 \\
Short-course & SSM & 0.765 $\pm$ 0.004 & 0.743 $\pm$ 0.003 & 0.441 $\pm$ 0.008 & 0.448 $\pm$ 0.003 & 0.381 $\pm$ 0.008 & 0.356 $\pm$ 0.005 \\
\bottomrule
\end{tabular}

}
\end{table}

\clearpage
\subsection{Within-admission correlation of performance metrics}
\label{app:within_admission_correlation_performance_metrics}

\begin{table}[h!]
    \centering
    \scriptsize
    \caption{Comparing AUROCs of models across targets for naive to clustered bootstrapping of test patient days in \emph{PIC}.}
    \label{tab:within_admission_correlation_performance_metrics}
  \resizebox{\textwidth}{!}{\begin{tabular}{llll}
    \toprule
    Target & Model & AUROC [cluster 95\% CI] & AUROC [naive 95\% CI] \\
    \midrule
    IV-to-oral & Logistic regression & 0.934 [0.892--0.971] & 0.934 [0.892--0.968] \\
    IV-to-oral & LightGBM & 0.932 [0.897--0.959] & 0.932 [0.901--0.961] \\
    IV-to-oral & MLP & 0.926 [0.873--0.972] & 0.926 [0.877--0.970] \\
    IV-to-oral & GRU & 0.956 [0.925--0.980] & 0.956 [0.929--0.979] \\
    IV-to-oral & LSTM & 0.938 [0.893--0.970] & 0.938 [0.895--0.974] \\
    IV-to-oral & TCN & 0.950 [0.913--0.977] & 0.950 [0.917--0.977] \\
    IV-to-oral & Transformer & 0.950 [0.916--0.976] & 0.950 [0.917--0.976] \\
    IV-to-oral & SSM & 0.954 [0.924--0.977] & 0.954 [0.925--0.976] \\
    \addlinespace
    De-escalation & Logistic regression & 0.911 [0.899--0.923] & 0.911 [0.901--0.920] \\
    De-escalation & LightGBM & 0.936 [0.928--0.945] & 0.936 [0.929--0.942] \\
    De-escalation & MLP & 0.920 [0.908--0.931] & 0.920 [0.910--0.928] \\
    De-escalation & GRU & 0.964 [0.957--0.969] & 0.964 [0.959--0.971] \\
    De-escalation & LSTM & 0.963 [0.957--0.970] & 0.963 [0.957--0.968] \\
    De-escalation & TCN & 0.964 [0.957--0.970] & 0.964 [0.958--0.969] \\
    De-escalation & Transformer & 0.953 [0.946--0.959] & 0.953 [0.947--0.957] \\
    De-escalation & SSM & 0.959 [0.953--0.965] & 0.959 [0.954--0.964] \\
    \addlinespace
    Discontinuation & Logistic regression & 0.799 [0.777--0.821] & 0.799 [0.775--0.821] \\
    Discontinuation & LightGBM & 0.842 [0.823--0.860] & 0.842 [0.822--0.862] \\
    Discontinuation & MLP & 0.734 [0.708--0.760] & 0.734 [0.708--0.759] \\
    Discontinuation & GRU & 0.841 [0.821--0.860] & 0.841 [0.821--0.862] \\
    Discontinuation & LSTM & 0.843 [0.821--0.863] & 0.843 [0.823--0.863] \\
    Discontinuation & TCN & 0.837 [0.817--0.858] & 0.837 [0.817--0.855] \\
    Discontinuation & Transformer & 0.829 [0.805--0.851] & 0.829 [0.809--0.848] \\
    Discontinuation & SSM & 0.815 [0.797--0.837] & 0.815 [0.792--0.838] \\
    \addlinespace
    Short course & Logistic regression & 0.746 [0.729--0.763] & 0.746 [0.728--0.762] \\
    Short course & LightGBM & 0.805 [0.788--0.820] & 0.805 [0.787--0.819] \\
    Short course & MLP & 0.761 [0.744--0.782] & 0.761 [0.744--0.779] \\
    Short course & GRU & 0.858 [0.845--0.871] & 0.858 [0.847--0.869] \\
    Short course & LSTM & 0.844 [0.830--0.858] & 0.844 [0.830--0.857] \\
    Short course & TCN & 0.843 [0.829--0.856] & 0.843 [0.829--0.856] \\
    Short course & Transformer & 0.818 [0.803--0.833] & 0.818 [0.804--0.832] \\
    Short course & SSM & 0.784 [0.768--0.801] & 0.784 [0.769--0.802] \\
    \bottomrule
\end{tabular}}
\end{table}

\end{document}